\pdfoutput=1
\documentclass[11pt]{article}

\usepackage[final]{acl}

\usepackage{times}
\usepackage{latexsym}

\usepackage{graphicx}
\usepackage{subcaption}

\usepackage[T1]{fontenc}
\usepackage[utf8]{inputenc}

\usepackage{microtype}

\usepackage{inconsolata}

\usepackage{graphicx}
\usepackage{amsmath}
\usepackage{tikz}
\usepackage{multirow}
\usepackage{tcolorbox}
\def\checkmark{\tikz\fill[scale=0.4](0,.35) -- (.25,0) -- (1,.7) -- (.25,.15) -- cycle;} 

\newtcolorbox{graybox}{
  colback=gray!20,  % Background color (10% gray)
  boxrule=0.4pt,      % No border
  arc=5pt,          % No rounded corners
  boxsep=5pt,       % Internal padding
  left=3pt,        % Left margin
  right=3pt,       % Right margin
  top=3pt,          % Top margin
  bottom=3pt        % Bottom margin
}

\title{Decoding Knowledge in Large Language Models: A Framework for Categorization and Comprehension}

\author{Yanbo Fang \\
  \texttt{yanbofang0822@gmail.com} \\\And
  Ruixiang Tang \\
  Rutgers University \\
  \texttt{ruixiang.tang@rutgers.edu} \\}

\begin{document}
\maketitle
\begin{abstract}
Understanding how large language models (LLMs) acquire, retain, and apply knowledge remains an open challenge. This paper introduces a novel framework, K-(CSA)², which categorizes LLM knowledge along two dimensions: correctness and confidence. The framework defines six categories of knowledge, ranging from highly confident correctness to confidently held misconceptions, enabling a nuanced evaluation of model comprehension beyond binary accuracy. Using this framework, we demonstrate how techniques like chain-of-thought prompting and reinforcement learning with human feedback fundamentally alter the knowledge structures of internal (pre-trained) and external (context-dependent) knowledge in LLMs. CoT particularly enhances base model performance and shows synergistic benefits when applied to aligned LLMs. Moreover, our layer-wise analysis reveals that higher layers in LLMs encode more high-confidence knowledge, while low-confidence knowledge tends to emerge in middle-to-lower layers. %These findings provide deeper insights into the hierarchical organization of knowledge within LLMs and the transformative impact of advanced prompting and alignment techniques.
%This framework offers unique insights into model behavior, suggesting new directions for improving knowledge integration in LLMs and providing quantitative metrics for assessing model development.

% Understanding how large language models (LLMs) acquire, retain, and apply knowledge remains an open challenge. This paper introduces a novel framework, K-(CSA)², which categorizes LLM knowledge along two dimensions: correctness and confidence. The framework defines six categories of knowledge, ranging from highly confident correctness to confidently held misconceptions, enabling a nuanced evaluation of model comprehension beyond binary accuracy. Using this framework, we analyze how techniques like chain-of-thought prompting and reinforcement learning with human feedback influence the evolution of internal (pre-trained) and external (context-dependent) knowledge in LLMs. Experimental results on multiple models reveal distinct patterns in knowledge stability, transitions, and integration, offering insights into optimizing training and evaluation strategies. Our findings provide a comprehensive lens for understanding and improving LLMs’ capacity for reliable knowledge representation and reasoning.

% Moreover, using our framework, We can trace how knowledge comprehension evolves through processes such as CoT and RLHF. \textcolor{red}{need a good findings from our paper here} The proposed method offers a new, fine-grained approach to understanding the knowledge learned by these models and may provide a novel metric for future evaluations.

\end{abstract}

\section{Introduction}

LLMs have demonstrated exceptional capabilities across various tasks, often matching or exceeding human performance \cite{surpasshuman1, surpasshuman2, surpasshuman3}. However, our understanding of how these models acquire and utilize knowledge remains limited \cite{Gekhman2024DoesFL, llmknowledge1, llmknowledge2}. Despite recent investigations \cite{framework1, framework2, framework3, framework4}, there lacks a comprehensive framework for categorizing and analyzing LLM knowledge types, particularly in relation to advanced techniques like chain-of-thought prompting \cite{wei2022chain}, instruction tuning \cite{instruction-tuning}, and RLHF \cite{rlhf}. This paper introduces such a framework and examines how these techniques influence LLM knowledge representation.

% We can change may confident unknown to mixed-confident unknown, or partial-confident unknown.

\begin{table}[t]
    \small
    \centering
    \begin{tabular}{|l|c|l|}
        \hline
        \multicolumn{3}{|l|}{\textbf{Question:} What is the largest planet in our solar system?} \\
        \multicolumn{3}{|l|}{\textbf{Answer:} Jupiter} \\
        \hline
        \textbf{Category} & \textbf{Greedy} & \textbf{Sampled} \\
        \hline
        \textbf{1. HK} & \colorbox{green!30}{Jupiter} & [\colorbox{green!30}{Jupiter}, \colorbox{green!30}{Jupiter}, ..., \colorbox{green!30}{Jupiter}] \\
        \textbf{2. MK} & \colorbox{green!30}{Jupiter} & [Mars, Saturn, ..., Mars] \\
        \textbf{3. WK} & Saturn & [Saturn, \colorbox{green!30}{Jupiter}, ..., Saturn] \\
        \textbf{4. UU} & Saturn & [Mars, Venus, ..., Earth] \\
        \textbf{5. MU} & \colorbox{red!30}{Saturn} & [\colorbox{red!30}{Saturn}, Mars, ..., \colorbox{red!30}{Saturn}] \\
        \textbf{6. CU} & \colorbox{red!30}{Saturn} & [\colorbox{red!30}{Saturn}, \colorbox{red!30}{Saturn}, ..., \colorbox{red!30}{Saturn}] \\
        \hline
    \end{tabular}
    \caption{Model example responses across different categories using Greedy (deterministic) and Sampled (stochastic) decoding strategies. We collect one response using greedy decoding and multiple responses using sampling. The \colorbox{green!30}{green} highlight color in the table indicates correct answers (Jupiter), while for duplicated incorrect answers (e.g., repeated Saturn), we use \colorbox{red!30}{red} highlight. Here are the full names of categories: \textbf{1. HK}: highly known, \textbf{2. MK}: maybe known, \textbf{3. WK}: weakly known, \textbf{4. UU}: unconfident unknown, \textbf{5. MU}: may confident unknown, \textbf{6. CU}: confident unknown.}
    \label{table:model-responses}
\end{table}

\begin{figure*}[hbt!]
    \centering
    \includegraphics[width=1.0\linewidth, height=6.5cm]{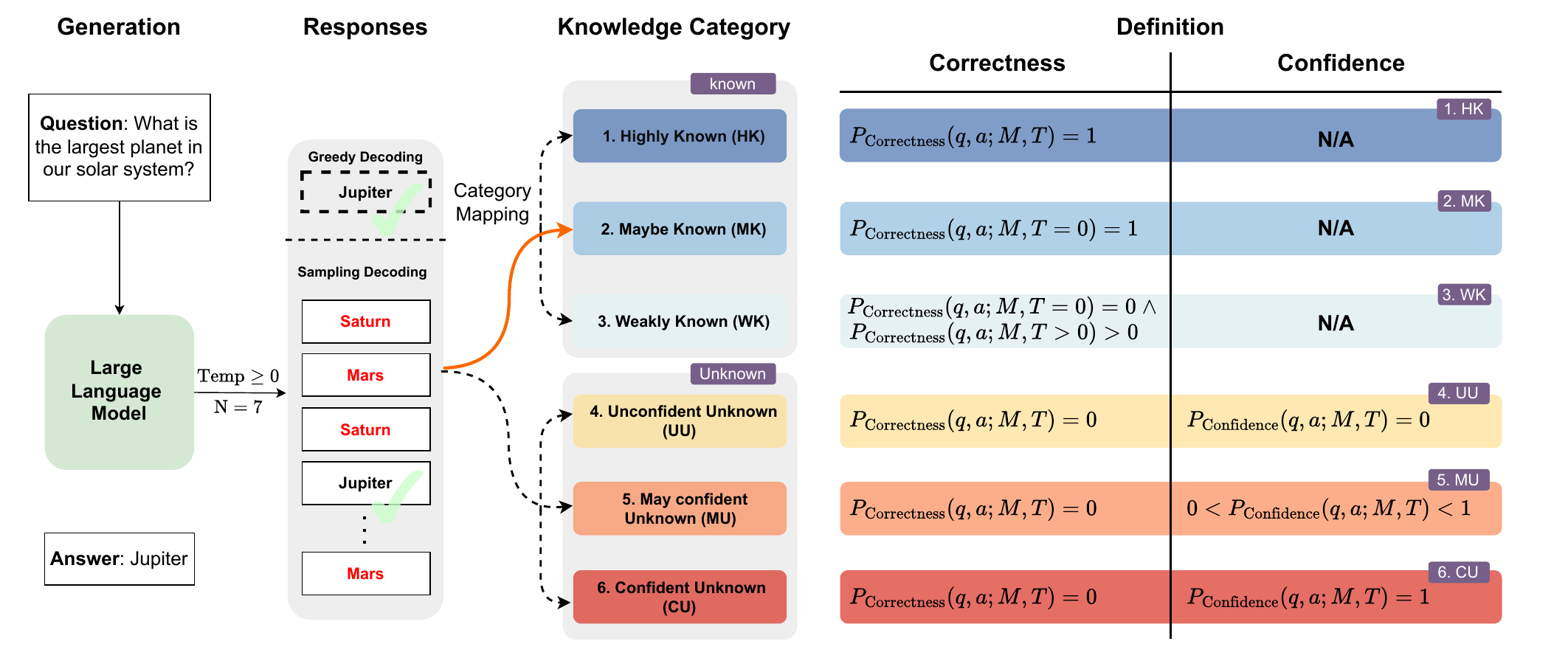}
    \caption{Illustration of our framework, K-(CSA)$^2$, for categorizing knowledge comprehension in LLMs.
The framework separates model responses into six categories: three for known knowledge (Highly Known (HK), Maybe Known (MK), Weakly Known (WK)) and three for unknown knowledge (Unconfident Unknown (UU), Mayconfident Unknown (MU), Confident Unknown (CU)). Greedy decoding represents deterministic answers, while random sampling introduces variability. Categories are defined based on both correctness and confidence in the model's responses, with responses mapped to categories based on confidence and correctness. The language model is represented by $M$, while $T$ represents the temperature. $q$ is the input question and $a$ is the LLM answer for $q$. The correctness probability is denoted by $P_{\textrm{Correctness}}$, and the confidence level is indicated by $P_{\textrm{Confidence}}$.}
    \label{fig:definition}
\end{figure*}

Current evaluations of large language models focus primarily on answer correctness, overlooking the crucial aspects of confidence and consistency in knowledge representation \cite{framework2}. A model generating a correct answer through uncertain sampling differs fundamentally from one producing it consistently, yet traditional metrics fail to capture this distinction. We propose a knowledge categorization framework that examines both correctness and confidence, revealing deeper insights into how models acquire and express knowledge. We describe Related work in Section \ref{sec:related_work}.

% Existing evaluation approaches primarily focus on correctness—assessing whether an LLM can provide accurate answers based on inference. While this is important, such methods fail to account for how confident or consistent the model is in its predictions, which is equally crucial for understanding its knowledge capacity. In this paper, we propose a new knowledge categorization  framework that expands beyond correctness to incorporate model confidence. Specifically, we evaluate not just whether the model knows the right answer, but also how stable and confident it is in that knowledge. 

Our framework categorizes an LLM’s knowledge using two key dimensions: correctness and confidence. Correctness is assessed by comparing the model’s responses to ground-truth answers, while confidence is determined by repeatedly querying the model with variations of the same question. If the model consistently returns a stable and coherent answer across these variations, we infer that it holds its knowledge with greater confidence. By integrating these two dimensions, our framework offers a more nuanced evaluation, capturing not only whether LLMs are accurate but also how strongly they “believe” in their answers.

We classify LLM knowledge into two primary types (Known and Unknown), each further divided into three categories. This categorization  scheme is summarized in Table \ref{tab:knowledge_category_description} and illustrated in Table \ref{table:model-responses} and Figure \ref{fig:definition}. We refer to this framework as \textbf{K-(CSA)$^2$} (\textbf{K}nowledge Categorization using \textbf{C}orrectnes\textbf{S} \textbf{A}nd \textbf{C}onfidence via \textbf{SA}mpling). Table \ref{table:model-responses} illustrates how this framework categorizes model responses for an example question. Our contributions are summarized as follows:

% Specifically, the categories are:\newline
%  (1) \textbf{Type: Known} – 1. \textit{Highly Known} (HK), 2. \textit{Maybe Known} (MK), 3. \textit{Weakly Known} (WK).\newline
%  (2) \textbf{Type: Unknown} – 4. \textit{Unconfident Unknown} (UU), 5. \textit{MayConfident Unknown} (MU), 6. \textit{Confident Unknown} (CU).

%Figure 1 demonstrates how incorporating contextual knowledge can improve a single model’s knowledge quality over time, while Figure 1b illustrates how different knowledge categories shift as training progresses.

\begin{itemize}
\item \textbf{Knowledge Categorization Framework}: We introduce K-(CSA)² framework to evaluate LLMs' knowledge by considering both correctness and confidence, providing a scalable approach for assessing model consistency across different tasks.
\item \textbf{Comprehensive Knowledge Analysis}: We examine how LLMs utilize both internal knowledge (from pre-trained weights) and external knowledge (from context), revealing distinct patterns in how models leverage these different knowledge types.
\item \textbf{Multi-Dimensional Knowledge Evolution}: Our experiments show that internal knowledge benefits more from CoT prompting, while external knowledge improves through instruction tuning, indicating different optimization strategies may be needed for different knowledge types.
    
%Through extensive experiments on both open-source and close-source LLMs, the study displays how knowledge categories shift over time and discussing the knowledge improving performance and features with specific methods like CoT prompting and RLHF.

    % \item \textbf{Analysis of }
\end{itemize}

%These category numbers $c$ are ordered from 1 to 6, followed by category name, where Category 1(HK) represents mastered knowledge, while Category 6(CU) denotes the least comprehended knowledge. Further details about this framework are provided in Section PLACEHOLDER, with specific definitions of each category illustrated in Figure 2.

\section{Framework and Metric Development for Knowledge Analysis}
\subsection{Knowledge Categorization Framework}
We evaluate large language models' knowledge comprehension using a six-category framework (Figure~\ref{fig:definition}) that distinguishes between \textit{known} and \textit{unknown} knowledge based on the model's ability to answer questions correctly. A knowledge point is defined as a specific piece of information required to correctly answer a given question (e.g., \textit{"Jupiter is the largest planet in our solar system"}). A knowledge point is classified as \textit{known} if the model produces the correct answer at least once across multiple sampling attempts. Conversely, it is categorized as \textit{unknown} if the model consistently generates incorrect responses.

Building on prior work \cite{Gekhman2024DoesFL}, our framework ranks knowledge comprehension from the most comprehensible category, \textit{Highly Known} (HK), to the least comprehensible, \textit{Confidently Unknown} (CU). For \textit{unknown} knowledge categories—Unconfident Unknown (UU), May Confident Unknown (MU), and Confidently Unknown (CU)—we incorporate confidence metrics to identify patterns in incorrect responses. CU is ranked lowest among unknown categories because it represents consistent but incorrect answers, diverging from the expected variability when models are uncertain. To classify responses into these categories, we analyze both \textbf{Correctness} and \textbf{Confidence}. Given a question \( q \), context \( c \), ground truth \( y \), and LLM \( M \), we generate \( n \) responses \( a \): one using greedy decoding (\( T=0 \)) and \( n-1 \) using random sampling (\( T>0 \)).\footnote{For closed-source LLMs, temperature \( T=0.5 \), following prior research~\cite{Gekhman2024DoesFL}. For close-source models, \( T=1 \).} Correctness is determined using exact match scoring, where at least one correct response categorizes the knowledge as \textit{known}. For incorrect responses, confidence is measured by evaluating the consistency and similarity among sampled outputs. The calculation method detail is further illustrated in \ref{subsec:confidence}.

\begin{figure*}[t]
    \centering
    \includegraphics[width=1\linewidth]{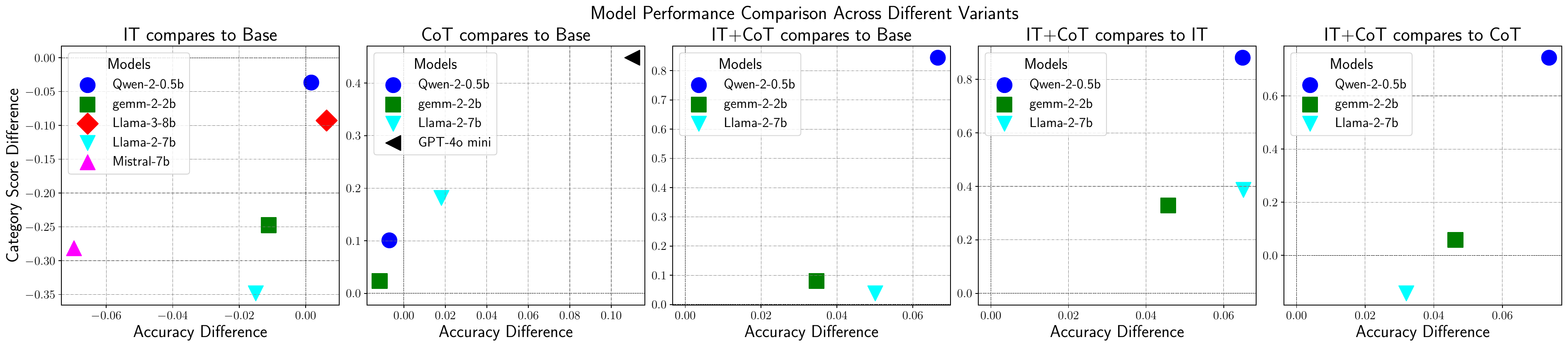}
    \caption{Comparative analysis of model \textbf{internal} knowledge performance across different variations. Each subplot demonstrates performance differences between model variants, measuring changes in both accuracy (x-axis) and category score (y-axis). From left to right: (1) IT versus base models, showing a general decrease in performance; (2) CoT versus base models, indicating moderate improvements; (3) IT+CoT versus base models, revealing substantial gains; (4) IT+CoT versus IT, highlighting the additive benefits of CoT; and (5) IT+CoT versus CoT, showing complementary effects of combining both techniques. The scattered points represent different models, with their relative positions indicating the magnitude and direction of performance changes. Positive values on both axes indicate improvement over the comparison baseline, while negative values suggest performance degradation. Base: Base model, IT: instructed model, CoT: chain-of-thought.}
    \label{fig:internal_scatter}
\end{figure*}

\subsection{Evaluation Metric}

\textbf{Evaluating Correctness and Confidence.}: We evaluate LLMs' correctness by accuracy. Using greedy decoding ($T=0$) against ground truth answers to assess their ability to generate accurate, coherent responses with minimal randomness.
While accuracy alone measures correctness, it fails to capture how confidently models hold their knowledge. To provide a more comprehensive evaluation of LLM knowledge comprehension, we introduce Category Scores as a complementary metric that integrates both correctness and confidence. While traditional accuracy metrics only capture whether responses are correct, our Category Score metric offers deeper insights by weighting responses according to their knowledge category classification. %This approach reflects the full spectrum of knowledge categories from highly confident correct answers to confidently held misconceptions, providing a more nuanced view of model performance. 
It is calculated as:
\begin{equation}
    \text{Category Score} = \sum_{i=1}^{6} w_i \cdot r_i
\end{equation}
where $w_i = 7-i$ represents the weight for category $i$ (from 6 for 1.HK to 1 for 6.CU), and $r_i$ is the ratio of responses falling into category $i$. The weighting scheme assigns higher values to more desirable categories, with 1.HK responses receiving the weight of 6 and 6.CU receiving the minimum weight of 1. This design penalizes models that consistently generate incorrect answers with high confidence, and it rewards models that express appropriate uncertainty when encountering unknown information, making it easier to identify genuine knowledge gaps. Examples are in section \ref{sec:example_category_score}.

\noindent \textbf{Dynamics of Knowledge Categories.} To analyze how LLMs' knowledge evolves, we track transitions of knowledge points (question-answer pairs) between different categories across model versions or training stages. A transition occurs when a model's handling of a specific knowledge point changes from one category to another. These transitions reveal distinct aspects of model behavior, captured through three complementary but mutually exclusive metrics that sum to 1\footnote{Some more explanations of transitions ratios are in section \ref{sec:explanaiton_ratios}.}:
\begin{itemize}
    \item \textbf{Upgrade Ratio}: The proportion of knowledge points that move to categories indicating better understanding (e.g., from 3.WK to 2.MK). A high upgrade ratio suggests effective learning to handle knowledge. This metric specifically tracks improvements.
    \item \textbf{Downgrade Ratio}: The proportion of knowledge points that shift to categories indicating deteriorated understanding (e.g., from 2.MK to 4.UU). A higher downgrade ratio may signal problematic changes in model behavior. This metric captures regressions in model.
    \item \textbf{Stable Ratio}: The proportion of knowledge points that maintain their category, indicating consistent model behavior. This helps distinguish between beneficial changes and potentially unstable learning patterns, measuring a different dimension of model transition than the upgrade or downgrade metrics.
    % \item \textbf{Upgrade Ratio}: The proportion of knowledge points that move to categories indicating better understanding (e.g., from 3.WK to 2.MK). A high upgrade ratio suggests effective learning to handle knowledge. This metric specifically tracks improvements in model comprehension, independent of stability or degradation patterns.
    % \item \textbf{Downgrade Ratio}: The proportion of knowledge points that shift to categories indicating deteriorated understanding (e.g., from 2.MK to 4.UU). A higher downgrade ratio may signal problematic changes in model behavior. This metric captures regressions in model performance, distinct from improvements or stability.

    % \item \textbf{Stable Ratio}: The proportion of knowledge points that maintain their category, indicating consistent model behavior. This helps distinguish between beneficial changes and potentially unstable learning patterns, measuring a different dimension of model evolution than the upgrade or downgrade metrics.
\end{itemize}
\begin{figure}[t]
    \centering
    \includegraphics[width=1.07\linewidth]{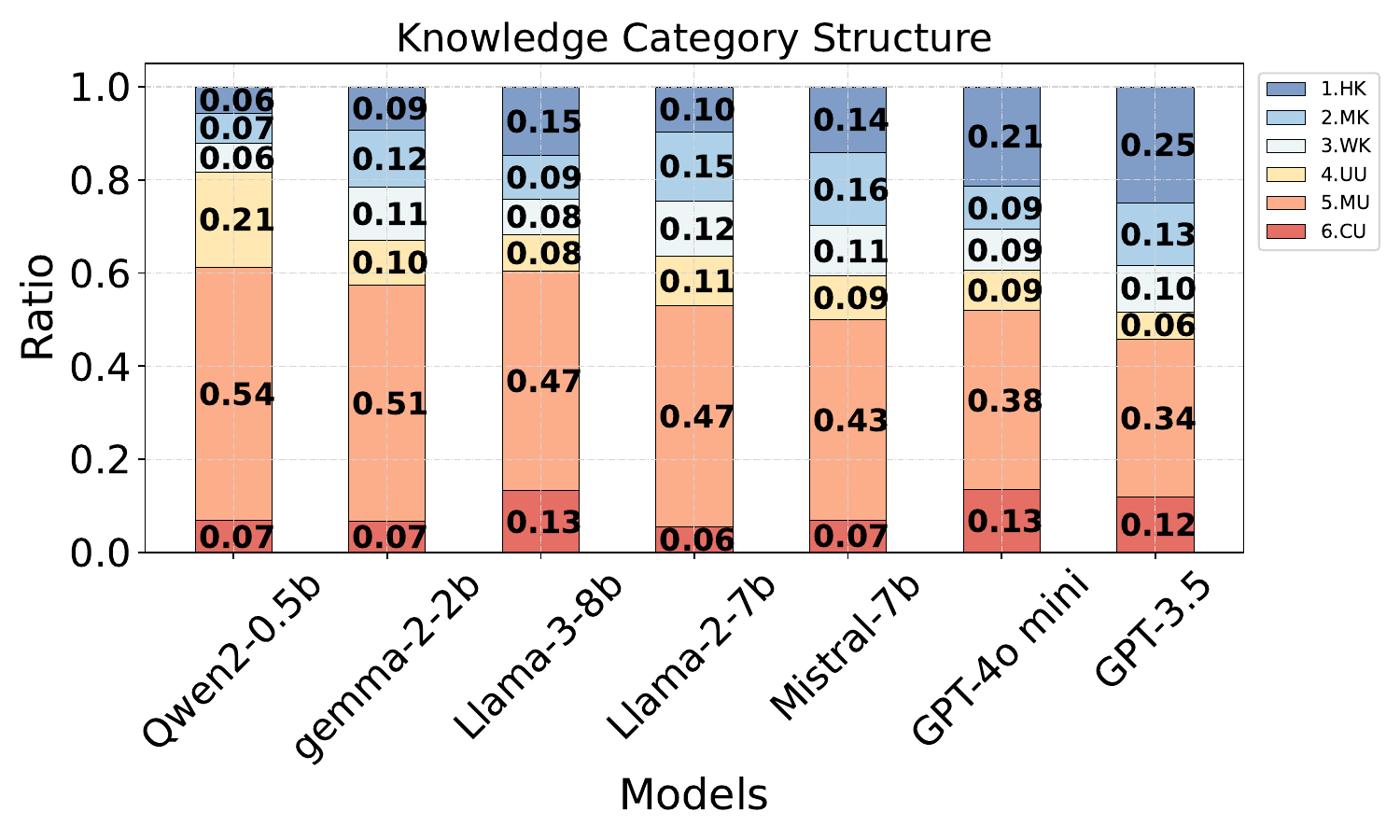}
    \caption{\textbf{Internal} knowledge categories structure across models, sorted left to right by increasing combined accuracy (ratios of top-2 layers 1.HK + 2.MK). The values within each section represent the ratio of knowledge points belonging to each category at different training steps. There are six knowledge categories, each represented by a different color.}
    \label{fig:internal_knowledge_structure}
\end{figure}

%[Original Performance's matter]

\section{Experiments}
% [Describing how we run our experiments, our selection of models, dataset, implementation detail]

In this section, we describe the experimental setup, the models used, and the datasets. With this framework (\textbf{K-(CSA)$^2$}), we designed our experiments to evaluate both the breadth of model capabilities and the evolution of their knowledge structures. For all experiments, we prompted the LLMs seven times for each question, using greedy decoding ($T=0$) once and random sampling ($T>0$) six times. This allows us to assess both correctness and confidence in the models’ knowledge. The key metrics analyzed include accuracy (via exact match with ground truth answers) and confidence across multiple generations. For the full list of LLMs and their capacity we evaluate, see appendix Table \ref{table:model_collection}.

%, The templates are also shown in [Appendix].

\textbf{Dataset.} We run experiments over the HaluEval dataset \cite{li-etal-2023-halueval}, which contains a knowledge-based QA task. In this dataset, one question $q$ pairs with an annotated knowledge point which can help answer the question and a ground-truth answer $a$. We randomly sampled $3000$ as our dataset, and we repeated prompting LLM with the same question $7$ times, once for greedy decoding and $6$ times randomly sampling ($T>0$). We run experiments with multiple closed-sourced and open-sourced LLMs.
\begin{figure*}[t]
    \centering
    \begin{subfigure}{0.492\linewidth}
        \centering
        \includegraphics[width=3.15in, height=1.22in]{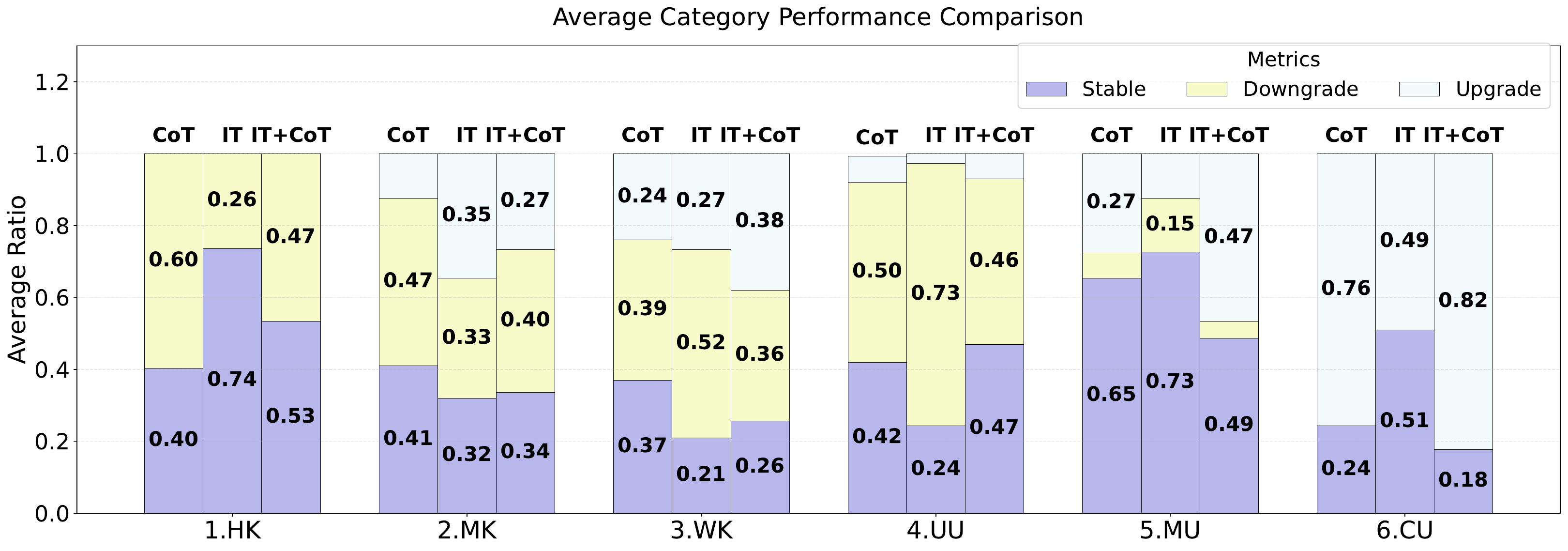}
        \caption{Internal Knowledge}
        \label{fig:internal_category_transition_it_category_mean}
    \end{subfigure}
    \begin{subfigure}{0.492\linewidth}
        \centering
        \includegraphics[width=3.15in, height=1.22in]{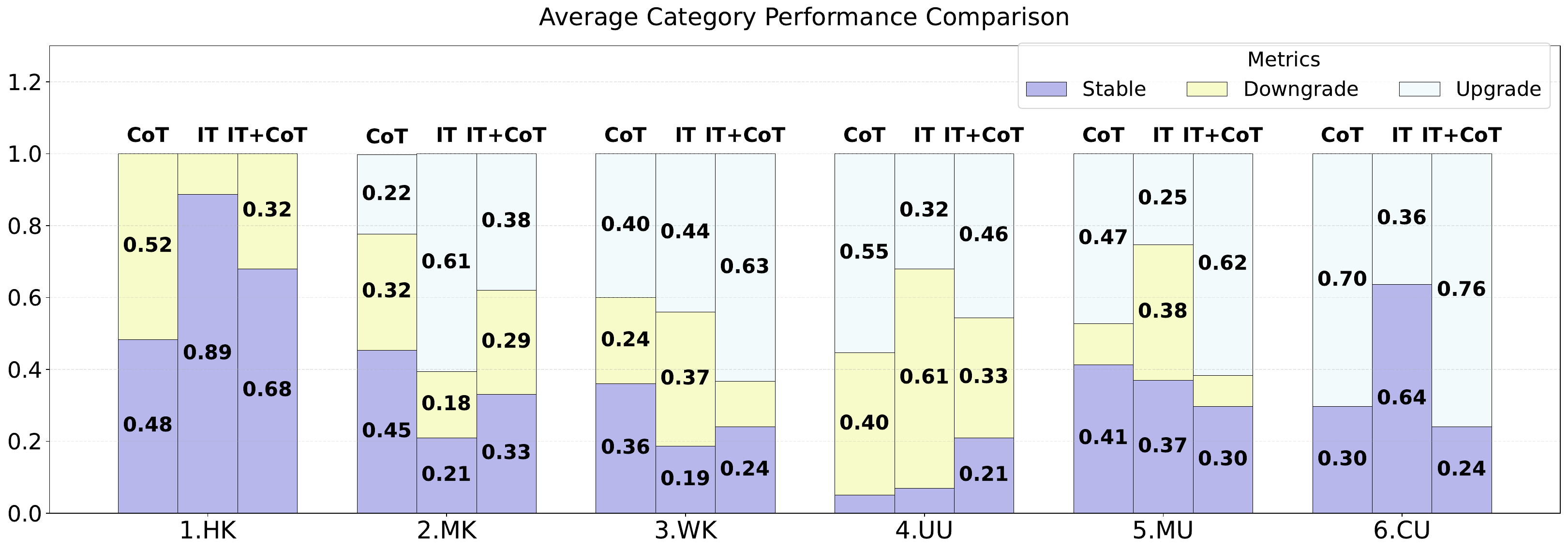}
        \caption{External Knowledge}
        \label{fig:external_category_transition_it_category_mean}
    \end{subfigure}

    \caption{Average transition patterns across all evaluated models, comparing how CoT and IT+CoT affect category transitions relative to the base model. Results are shown separately for (a) internal knowledge and (b) external knowledge, with each bar representing the mean stable, downgrade, and upgrade ratios for each knowledge category.}
    \label{fig:category_mean}
\end{figure*}

\subsection{Assess Internal Knowledge Utilization}

Internal knowledge refers to the information encoded within the LLM's weights during pre-training and fine-tuning phases, which enables the model to answer questions. We aim to use our proposed framework to illustrate the model's internal knowledge category structure and track how it evolves throughout pre-training, instruction-tuning, RLHF, and the application of techniques like CoT prompting, using the metrics and dataset mentioned above to understand this knowledge capacity. We prompt the model to answer given questions without annotated knowledge point.

%[Figures and Results and Analysis]

\begin{graybox}
    \textbf{Finding 1}: Stronger models are more assertive regardless of correctness.
\end{graybox}
We evaluate knowledge structures across a selection of open-source and proprietary models. As shown in Figure \ref{fig:internal_knowledge_structure}, we observe that models with higher accuracy tend to exhibit lower levels of uncertainty, suggesting that more capable LLMs are generally more certain about their responses. However, we also find that models can display confidence in incorrect answers. From left to right, the combined ratio of categories 3.WK, 4.UU, and 5.MU decreases, while accuracy-related categories 1.HK and 2.MK increase. Notably, the ratio of category 6.CU, representing the lowest understanding in our framework, also rises. This pattern raises considerations for model deployment, as higher capability doesn't necessarily guarantee better calibration between confidence and correctness. The simultaneous increase in both highly confident correct (1.HK) and incorrect (6.CU) responses suggests that model scaling may enhance assertiveness without proportionally improving the ability to recognize knowledge limitations.

% We use two metrics: \textbf{Accuracy} and \textbf{Category Score}. Because while Accuracy is a powerful metric to evaluate model's comprehension of knowledge, we show in our results Figure \ref{} 

% \begin{table*}[h!]
%     \centering
%     \begin{tabular}{r|ccccccc}
%         \hline
%         \textbf{Models} & \textbf{Qwen2-0.5b} & \textbf{Gemma-2b} & \textbf{Llama3-8b} & \textbf{Llama2-7b} & \textbf{Mistral-7b} & \textbf{GPT4omini} & \textbf{GPT-3.5} \\ \hline

%         \textbf{Internal} & 0.1210 & 0.2154 & 0.2407 & 0.2446 & 0.2983 & 0.3063 & 0.3830 \\

%         \textbf{External} & 0.4160 & 0.5800 & 0.5650 & 0.6076 & 0.7110 & 0.7514 & 0.7647 \\ \hline
%     \end{tabular}
%     \caption{Accuracy ($[0, 1]$) for internal knowledge Mastery (without context knowledge) and external knowledge understanding (with context knowledge)}
%     \label{table:accuracy}
% \end{table*}

% \begin{table*}[h!]
%     \centering
%     \begin{tabular}{r|ccccccc}
%         \hline
%         \textbf{Models} & \textbf{Qwen2-0.5b} & \textbf{Gemma-2b} & \textbf{Llama3-8b} & \textbf{Llama2-7b} & \textbf{Mistral-7b} & \textbf{GPT4omini} & \textbf{GPT-3.5} \\ \hline

%         \textbf{Internal} & 2.6798 & 2.9978 & 2.9706 & 3.1200 & 3.2749 & 3.2601 & 3.5371 \\

%         \textbf{External} & 3.6313 & 4.3014 & 4.1939 & 4.4028 & 4.7404 & 4.9101 & 4.9585 \\ \hline
%     \end{tabular}
%     \caption{Category Score ($[1, 6]$) for internal knowledge Mastery (without context knowledge) and external knowledge understanding (with context knowledge)}
%     \label{table:category_score}
% \end{table*}
\begin{figure}[t]
    \centering
    \begin{subfigure}{0.492\linewidth}
        \centering
        \includegraphics[width=3.92cm, height=2.3cm]{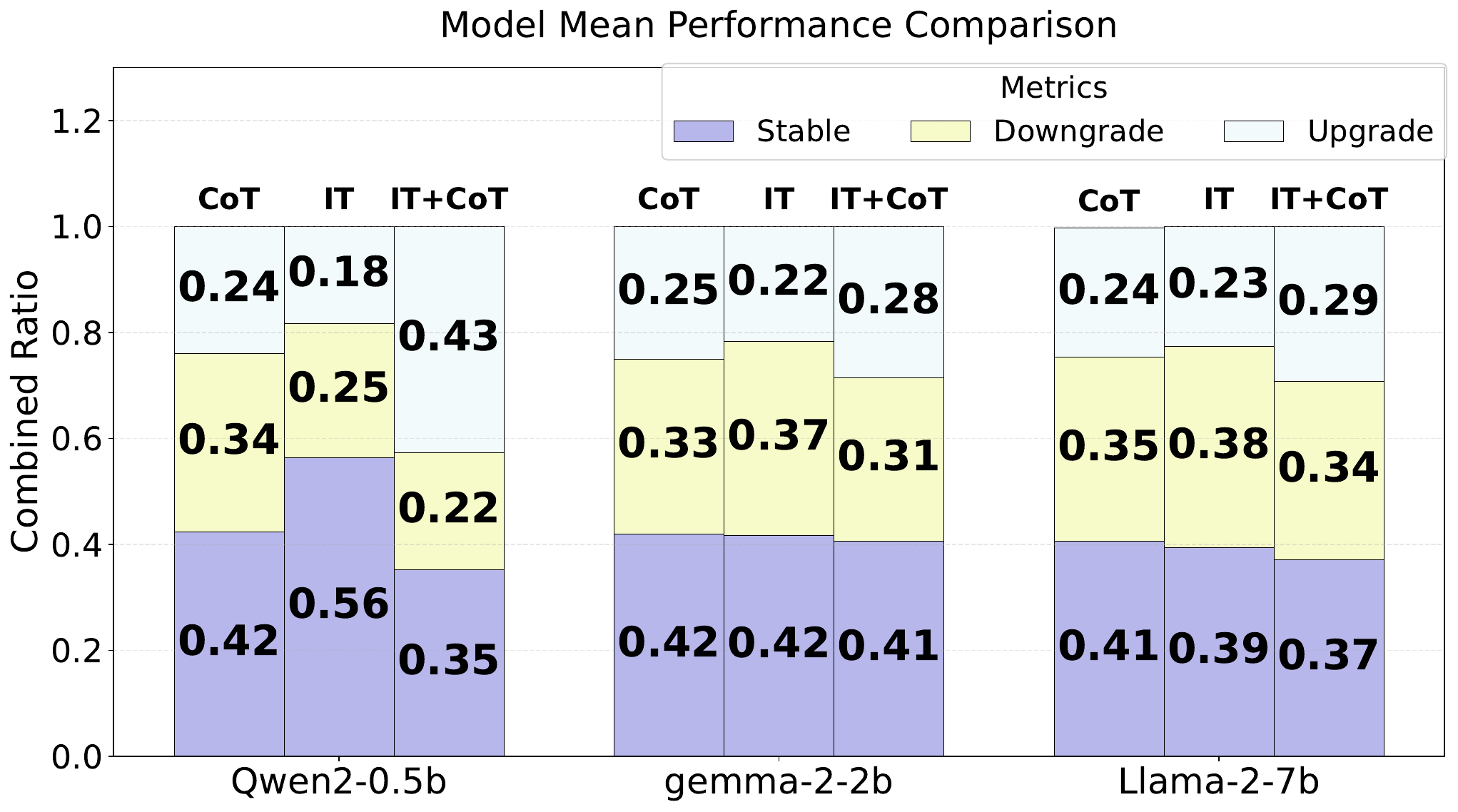}
        \caption{Internal Knowledge}
        \label{internal_category_transition_it_model_mean}
    \end{subfigure}
    \begin{subfigure}{0.492\linewidth}
        \centering
        \includegraphics[width=3.92cm, height=2.3cm]{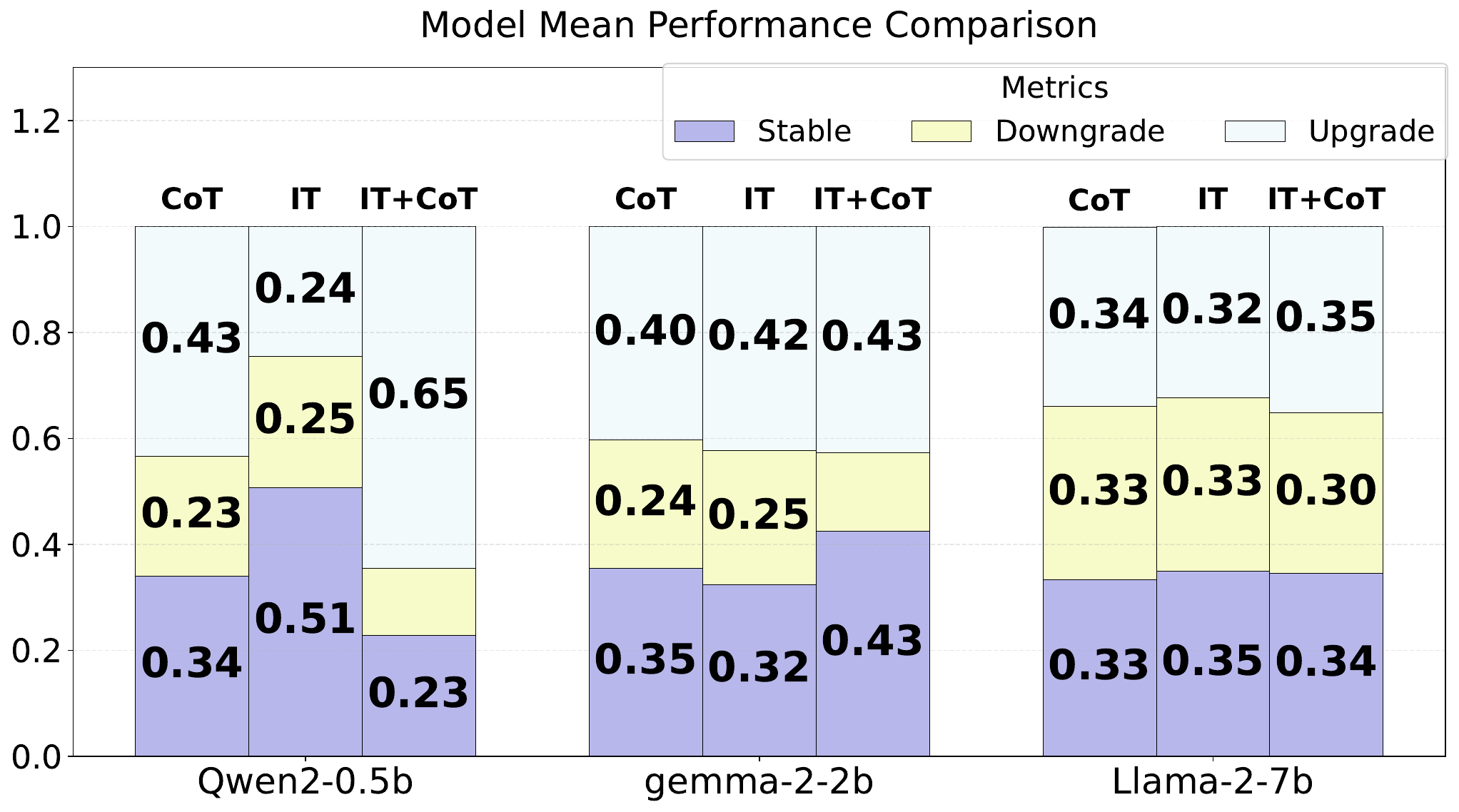}
        \caption{External Knowledge}
        \label{external_category_transition_it_model_mean}
    \end{subfigure}
    \caption{Mean category transition patterns for individual models, showing averaged effects of CoT and IT+CoT on (a) internal and (b) external knowledge performance relative to base models.}
    \label{fig:model_mean}
\end{figure}

\begin{graybox}
    \textbf{Finding 2}: CoT creates synergistic benefits when combined with instruction tuning despite possible initial performance drops from instruction tuning alone.
\end{graybox} 
We further examine how different techniques like CoT and instruction-tuning impact knowledge structures. These techniques fundamentally change how models process knowledge, and our framework reveals their effects across different training stages.
\begin{figure*}[t]
    \centering
    \includegraphics[width=1\linewidth]{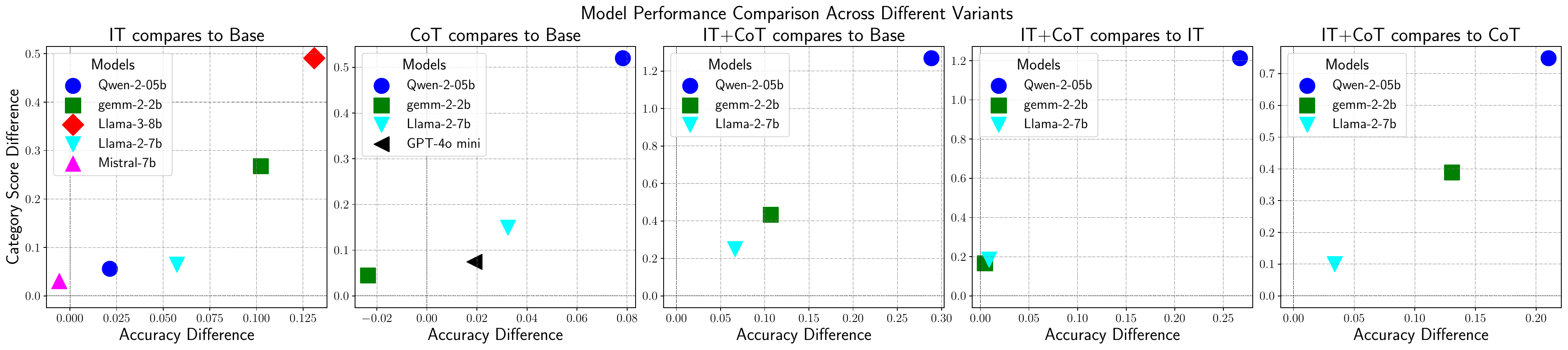}
    \caption{Comparative analysis of model \textbf{external} knowledge performance across different variations. Each subplot shows performance differences between variants in terms of accuracy (x-axis) and category score (y-axis). From left to right: (1) IT versus base models, showing varied improvements in both metrics; (2) CoT versus base models, demonstrating modest gains especially for Llama-2-7b; (3) IT+CoT versus base models, revealing significant improvements particularly in category scores; (4) IT+CoT versus IT, highlighting substantial additional benefits from combining with CoT; and (5) IT+CoT versus CoT, showing the complementary advantages of adding instruction tuning. Across all comparisons, Qwen-2-0.5b consistently shows strong improvements. \textbf{Base}: Base model, \textbf{IT}: instructed model, \textbf{CoT}: chain-of-thought.}
    \label{fig:external_scatter}
\end{figure*}
CoT enhances the base model's knowledge category structure and particularly benefits instructed models. Figure \ref{fig:internal_scatter} shows that applying CoT to base models increases category scores for models like Qwen2-0.5b, gemma-2-2b, GPT-4o mini, and Llama2-7b, though sometimes at the cost of accuracy. For example, gemma-2-2b and Qwen2-0.5b show slight accuracy decreases with CoT. For instructed models, CoT improves both accuracy and category scores (IT+CoT versus IT). Applying CoT to instructed versions of Qwen2-0.5b, gemma-2-2b, Llama2-7b, and GPT-4o mini enhances both metrics. While some models like Mistral-7b show initial deterioration after applying instruction tuning, CoT helps overcome this. Notably, while instruction tuning alone may not improve performance, it creates synergistic benefits when combined with CoT.
Our fine-grained analysis reveals that CoT enhances performance across all base models (Figure \ref{internal_category_transition_it_model_mean}). Combining IT with CoT increases upgrade ratios and decreasing downgrade ratios. These improvements primarily affect categories 3.WK through 6.CU, as shown in Figure \ref{fig:internal_category_transition_it_category_mean}, with detailed performance metrics presented in Figure \ref{internal_detailed} and More performance can be found in Figure \ref{fig:internal_category_transition_it_model_mean}.

\subsection{Assess External Knowledge Processing}

This section evaluates the capacity models that are used to understand external knowledge. This ability is particularly important for tasks such as open-book QA and retrieval-augmented generation. Our sampling-based framework can also be applied to evaluate the effectiveness of models integrating external knowledge. Similarly to internal knowledge evaluation, we analyze the evolution of external knowledge comprehension after instruction-tuning, and applying CoT. Different from evaluating internal knowledge, we attach knowledge points in context when prompting models for QA task.
%[Figures and Results and Analysis]
% \begin{graybox}
%     \textbf{Finding 3}: While external knowledge helps reduce model uncertainty, it doesn't always prevent confident errors, as shown by persistent high rates of CU responses.
% \end{graybox}

\begin{figure}
    \centering
    \includegraphics[width=1.07\linewidth]{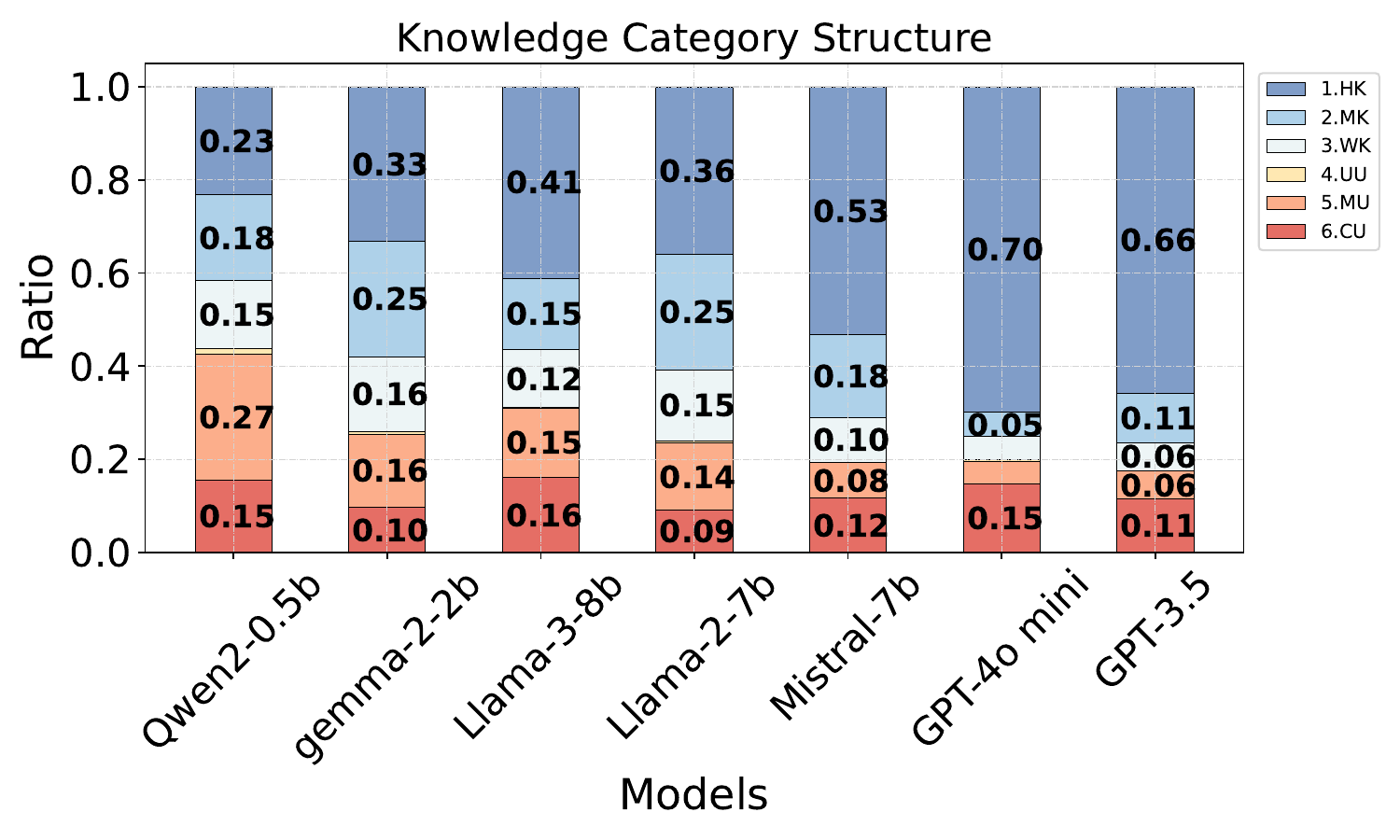}
    \caption{Breakdown of \textbf{external} knowledge category ratios across different language models, showing the relative proportions of each category from HK to CU.}
    \label{fig:external_knowledge_structure}
\end{figure}

The knowledge category structure for external knowledge is shown in Figure \ref{fig:external_knowledge_structure}. Similar to Figure \ref{fig:internal_knowledge_structure}, we sorted models from left to right based on their increasing accuracy values. By providing external knowledge, LLMs become more confident, reducing the proportions of categories 3.WK, 4.UU, and 5.MU. However, the reduction in these categories does not always translate to correct answers, as in category 6.CU remains high despite the addition of contextual knowledge.
\begin{figure*}[t]
    \centering
    \begin{subfigure}{0.493\linewidth}
        \centering
        \includegraphics[width=9.83cm, height=6cm]{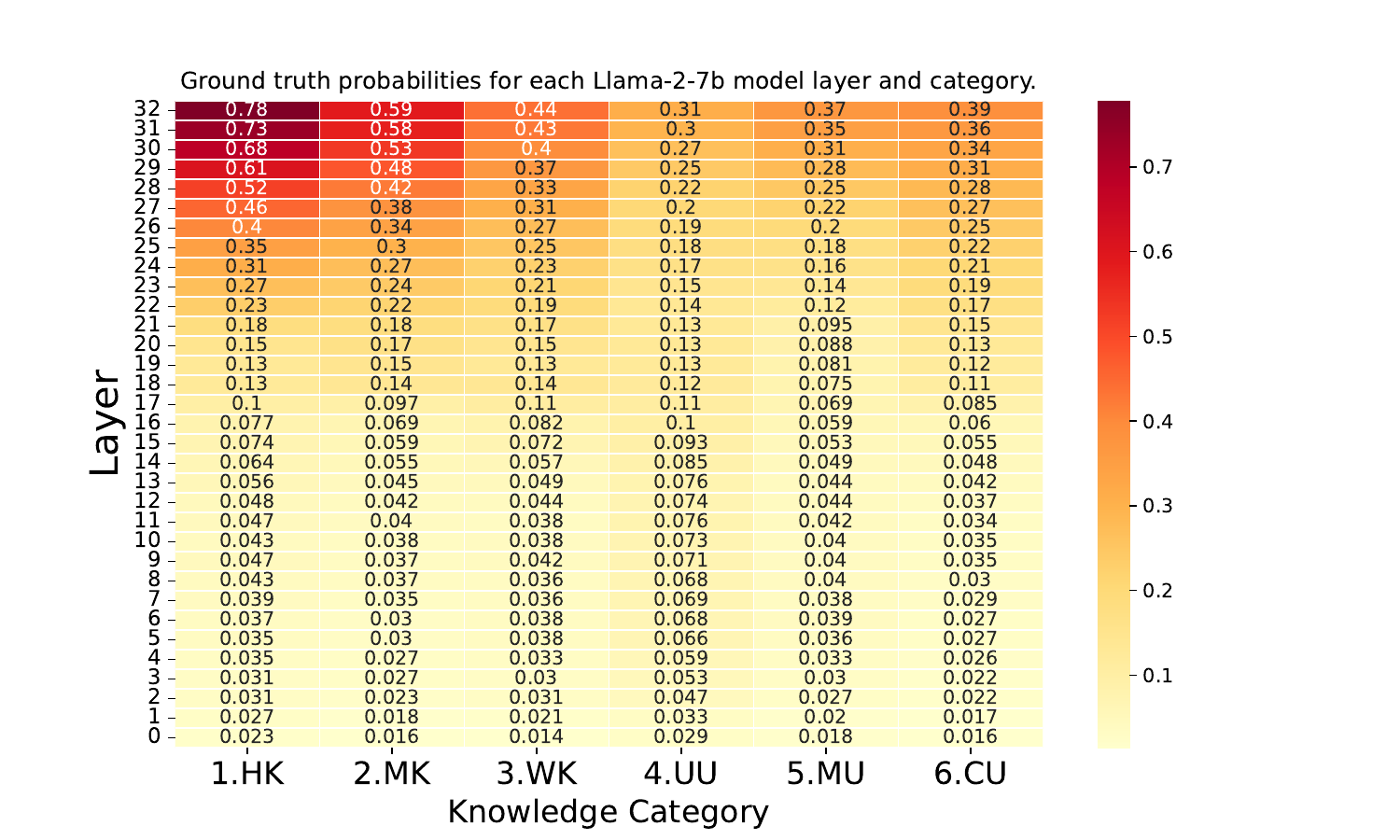}
        \caption{Llama-2-7b}
    \end{subfigure}
    \begin{subfigure}{0.493\linewidth}
        \centering
        \includegraphics[width=9.835cm, height=6cm]{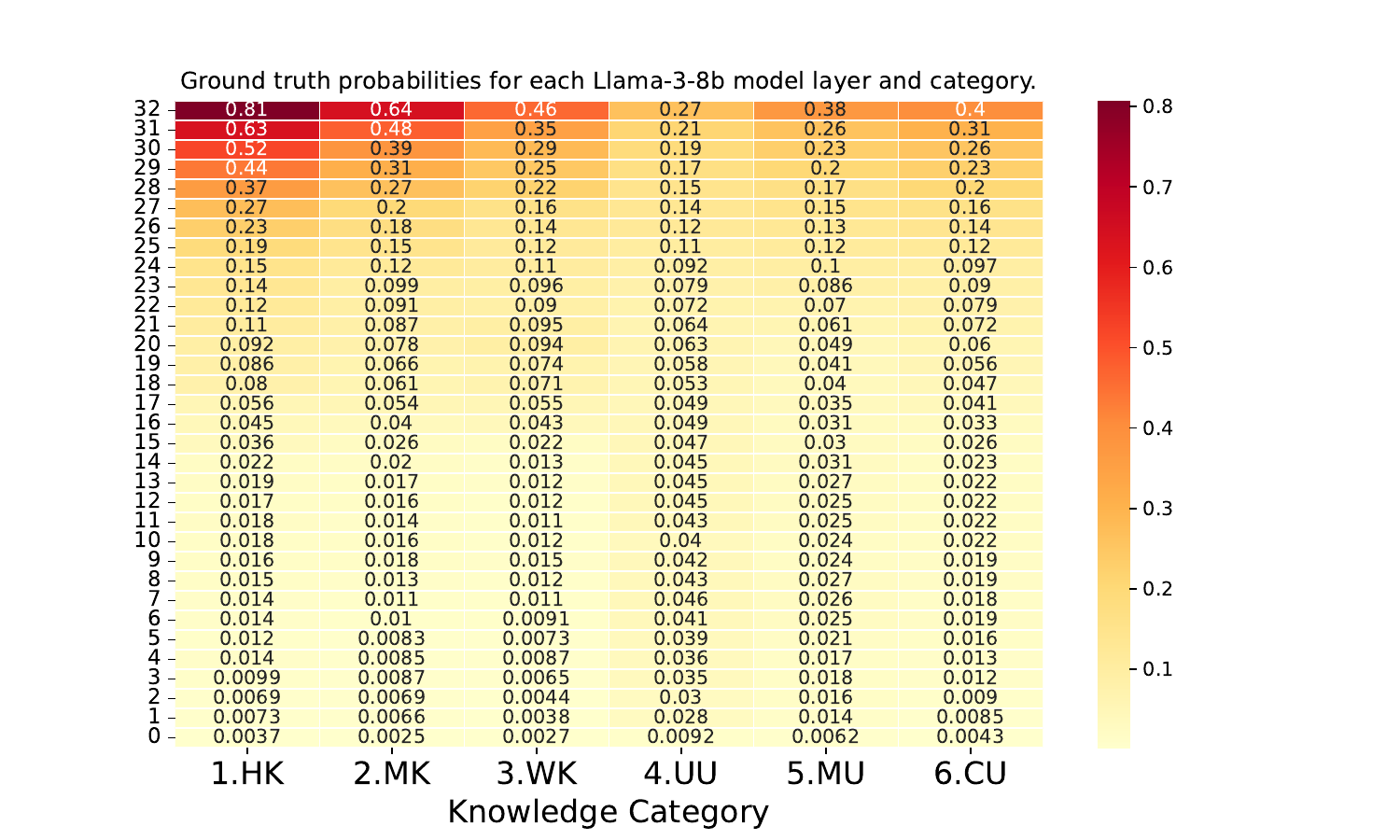}
        \caption{Llama-3-8b}
    \end{subfigure}
    \caption{Heatmap visualization of knowledge distribution across model layers for (a) Llama-2-7b and (b) Llama-3-8b. Each cell shows the probability of correct responses for a specific layer (y-axis) and knowledge category (x-axis), with darker red indicating higher probabilities.}
    \label{fig:lens}
\end{figure*}
\begin{graybox}
    \textbf{Finding 3}: Instruction-tuned models show superior context understanding, with IT+CoT providing more benefits for both internal and external knowledge processing.
\end{graybox} 
In Figure \ref{fig:external_scatter}, we analyze external knowledge patterns, finding that instructed models benefit more from contextual knowledge. This suggests that instructed LLMs are better at utilizing context for both understanding external information and generating CoT responses. While applying CoT to base models may not always improve accuracy (as seen with gemma2-2b), it typically enhances the overall knowledge category structure, indicating that both CoT and instruction tuning improves context utilization capacity. The combination of IT+CoT shows complementary benefits, as demonstrated in Figures \ref{fig:external_category_transition_it_category_mean} and \ref{external_category_transition_it_model_mean}. This combination increases upgrade ratios while reducing downgrade ratios, particularly affecting categories 3.WK through 6.CU. These patterns mirror our internal knowledge findings, suggesting consistent underlying mechanisms across different knowledge types (More detailed results are shown in Figure \ref{fig:others_model_performance} and \ref{external_detailed}).

\section{Further Analysis of Knowledge Category Distribution and Transition}

In this section, we examine knowledge organization in LLMs from two perspectives: spatial distribution across model layers (Section 4.1) and temporal development during training (Section 4.2). We analyze how different knowledge categories are distributed across network depths and how they evolve throughout the training process.

% \subsection{What if provided wrong context and unrelated context?}
% %[Investigate external knowledge and internal knowledge which has higher priority, and to what extent context knowledge would override internal knowledge]

\subsection{Layer Variations Across Categories}
%\textcolor{red}{motivation, why we need layer analysis}
Layer-wise analysis reveals important patterns about knowledge comprehension depth. Therefore, we analyze how model responses evolve across layers and categories through Figure \ref{fig:lens}, which shows layer-wise ground truth probabilities for Llama-3-8b and Llama-2-7b's focusing on model's internal knowledge. Ground truth probabilities are calculated by measuring the frequency of correct responses at each layer when the model was queried multiple times with the same input.
%\textbf{Higher layers in larger models comprehend more knowledge.} The heatmaps show ground truth probabilities increasing in higher layers %\textcolor{red}{how did you get the probability?}
\begin{graybox}
\textbf{Finding 4}: Higher layers in larger models encode more knowledge with greater intensity than smaller models.
\end{graybox}
Llama-3-8b's top layers (29-32) show consistently higher probabilities across most knowledge categories, with probabilities reaching 0.81 in layer 32 for HK category. The larger model allocates more capacity for knowledge representation in upper layers, while Llama-2-7b shows similar but weaker patterns, peaking at 0.78 in its highest layers. This layerwise distribution suggests that model size correlates with enhanced knowledge encoding capability in higher layers. The intensity difference is particularly pronounced in layers 25-32, where Llama-3-8b maintains higher probabilities across all knowledge categories, demonstrating more robust knowledge representation in its upper layers. This architectural difference indicates that larger models develop more pronounced hierarchical knowledge structures.
\begin{graybox}
\textbf{Finding 5}: uncertain knowledge emerges in lower layers but weakens in upper layers.
\end{graybox}
Categories with low confidence scores (4.UU and 5.MU) show higher ground truth probabilities in middle-to-lower layers (0-15). In both models, these categories exhibit distinct patterns: UU peaks around layers 10-15 with probabilities of 0.07-0.08, while MU shows slightly higher values around 0.1-0.12 in similar layers. While these categories are initially captured, they aren't strongly encoded in upper layers like 1.HK and 2.MK. Their probabilities increase gradually across lower layers before diminishing in upper layers, suggesting early-stage processing without deep integration. In Llama-3-8b, the pattern is particularly noticeable, with UU and MU probabilities declining more sharply after layer 20 compared to Llama-2-7b. This pattern is consistent across both model sizes, though more pronounced in Llama-2-7b where the contrast between lower and upper layer probabilities is greater. The weak uncertain knowledge probability in upper layers suggests that improving lower-confidence categories across upper layers could enhance the model's ability to generate correct answers and maintain proper uncertainty when needed.
\begin{graybox}
\textbf{Finding 6}: High confidence signals stronger knowledge comprehension in upper layers.
\end{graybox}
Both 1.HK and 6.CU categories show high ground truth probability in upper layers despite representing different types of understanding. In Llama-3-8b, these categories achieve probabilities of 0.81 and 0.31 respectively in the top layers, compared to Llama-2-7b's 0.78 and 0.39. This indicates that high confidence, regardless of correctness, corresponds to stronger knowledge encoding in top layers. The consistent pattern across both models reveals a fundamental aspect of how transformer architectures represent confident knowledge, with upper layers specializing in encoding high-confidence information regardless of its accuracy. This phenomenon suggests that the model's architecture inherently treats confident knowledge - whether correct or incorrect - similarly in its upper layers, potentially explaining why models can be simultaneously very confident and incorrect in their predictions. Similar patterns of correct and incorrect but confident knowledge suggests a need to distinguish between these types of knowledge.

% The similarity in encoding patterns between correct and incorrect but confident knowledge raises questions about how we might architecturally distinguish between these types of knowledge in future model designs.

\subsection{Knowledge Category Structure Transition During Pretraining}

Understanding knowledge accumulation and the capacity to comprehend context knowledge Evolution is essential for interpreting LLMs. In this section, we examine how models' knowledge category structures evolve during the pre-training process. We pick Pythia-2.8b and 1.4b to analyze both internal and external knowledge development patterns across different training stages.
%\textcolor{red}{add some description here}
\begin{figure}[t]
   \centering
   \includegraphics[width=8.2cm, height=3cm]{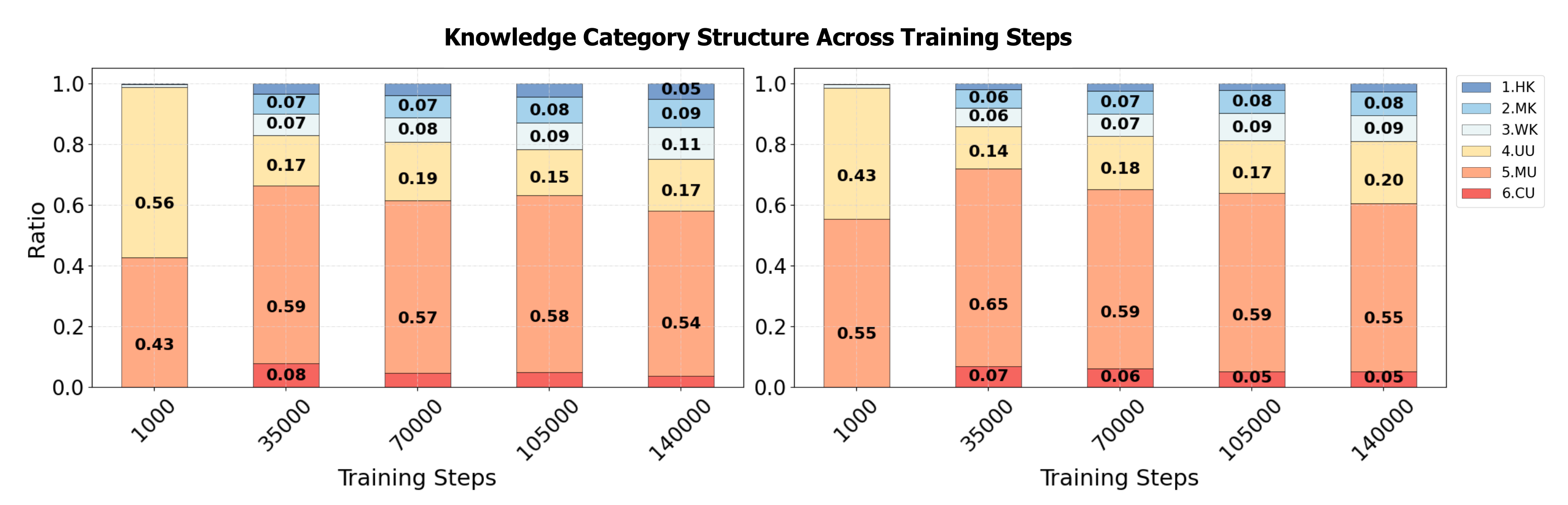}
   \caption{Changes in \textbf{internal} knowledge categories across different training steps for the \textbf{Pythia-2.8b} (\textbf{Left} figure) and \textbf{Pythia-1.4b} (\textbf{Right} figure) model}
    \label{fig:example_figure}
\end{figure}
\begin{figure}[]
    \centering
    \begin{subfigure}{0.493\linewidth}
        \centering
        \includegraphics[width=1.7in]{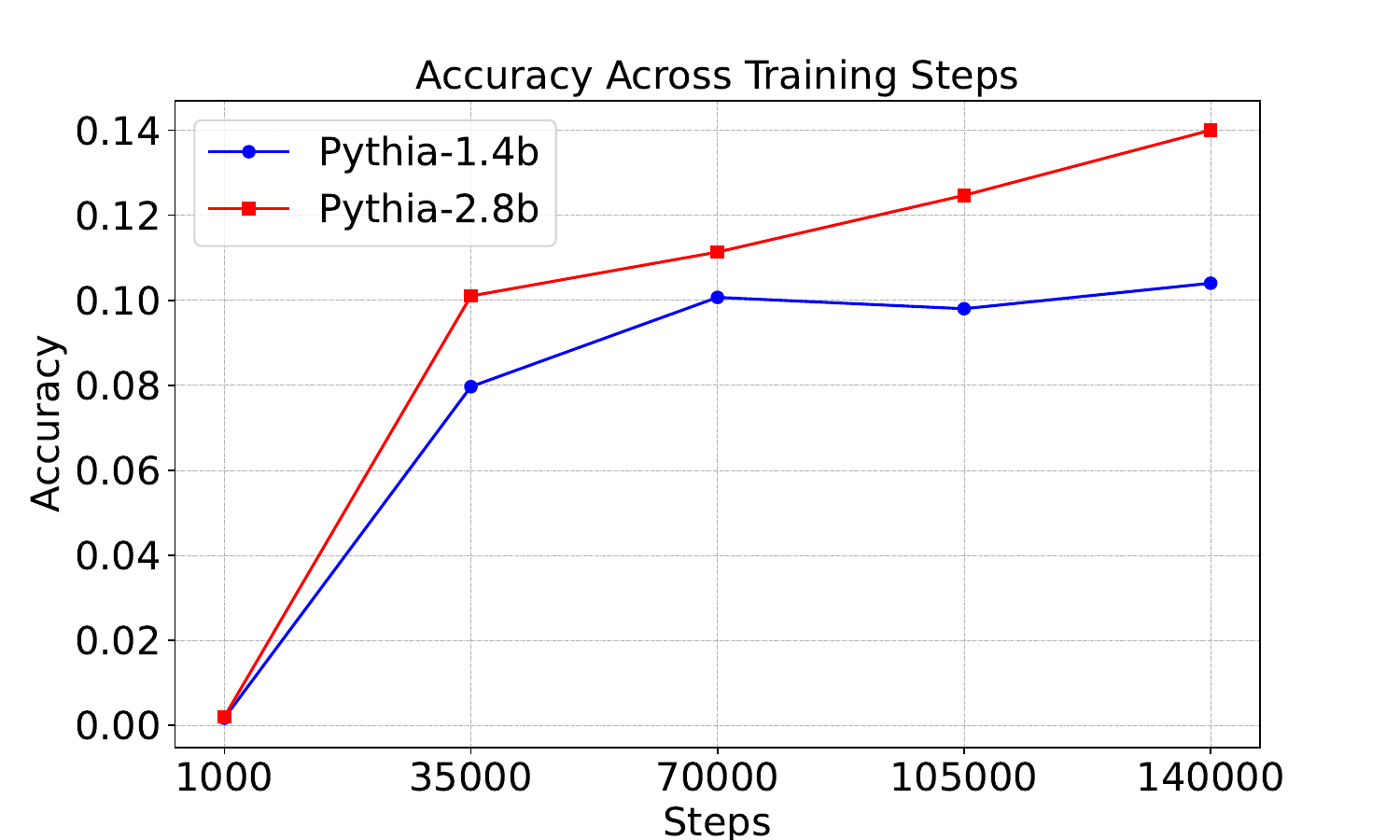}
        \caption{Accuracy}
    \end{subfigure}
    \begin{subfigure}{0.493\linewidth}
        \centering
        \includegraphics[width=1.7in]{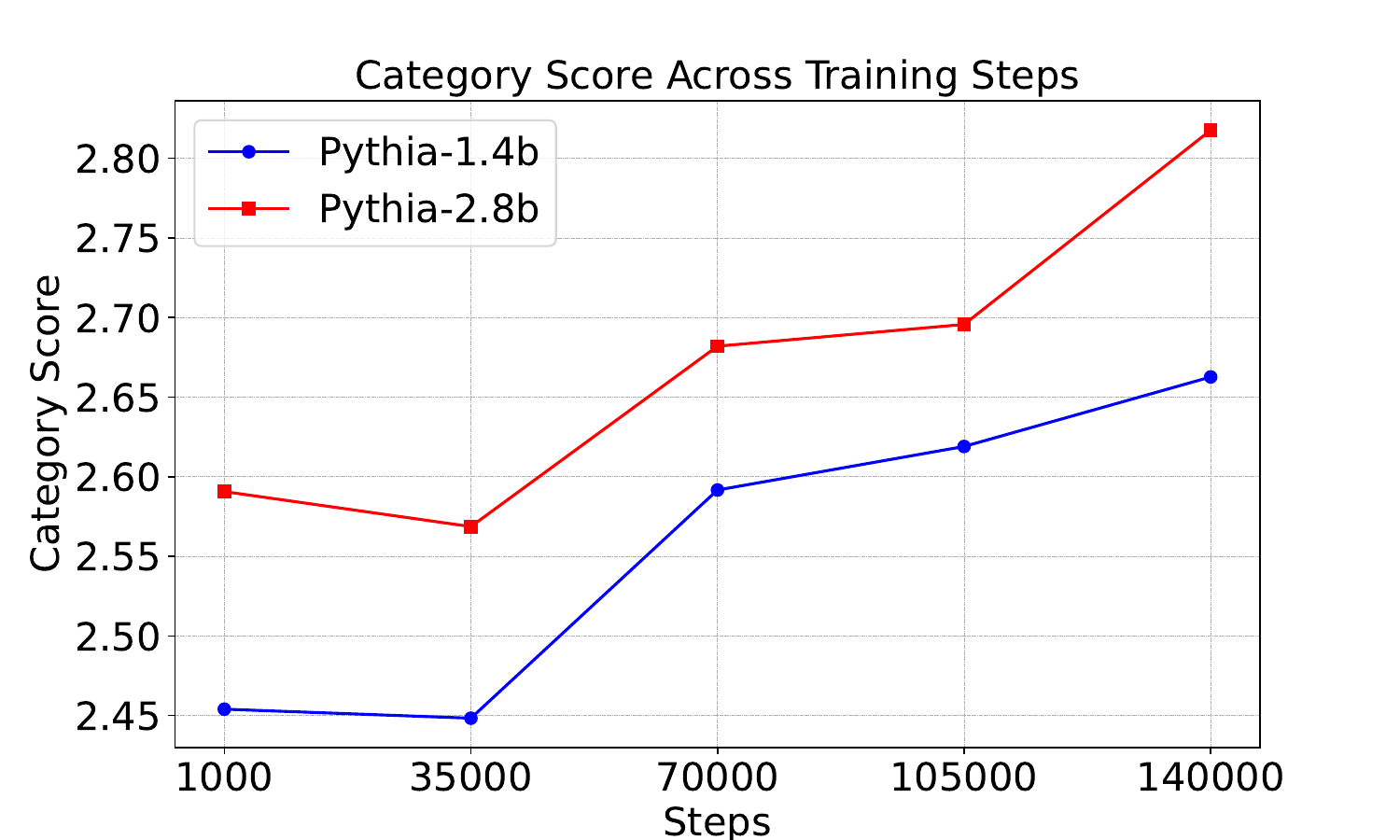}
        \caption{Category Score}
    \end{subfigure}
    \caption{Accuracy and Category Score for internal knowledge during different training steps.}
    \label{fig:internal_accuracy_and_category_score}
\end{figure}
\begin{graybox}
    \textbf{Finding 7}: Rising accuracy doesn't guarantee better knowledge structure.
\end{graybox} 
In Figure \ref{fig:example_figure}, we observe that the models transition from uncertainty to certainty throughout the training, as indicated by the decrease in the combined ratio, 4.UU+5.MU. However, this progression is not entirely smooth; for instance, the 5.MU category initially expands at around 35,000 steps before decreasing in subsequent training steps, even though accuracy continues to increase. Figure \ref{fig:internal_accuracy_and_category_score} illustrates changes in accuracy and category scores over training. The category score initially decreases before improving in later, while accuracy shows a steady upward trend. This indicates that training does not always consistently upgrade the structure of knowledge, and accuracy alone may not reliably reflect the depth of knowledge understanding. However, after additional iterations, accuracy and knowledge scores become more aligned.

\begin{figure}[t]
   \centering
   \includegraphics[width=8.2cm, height=3cm]{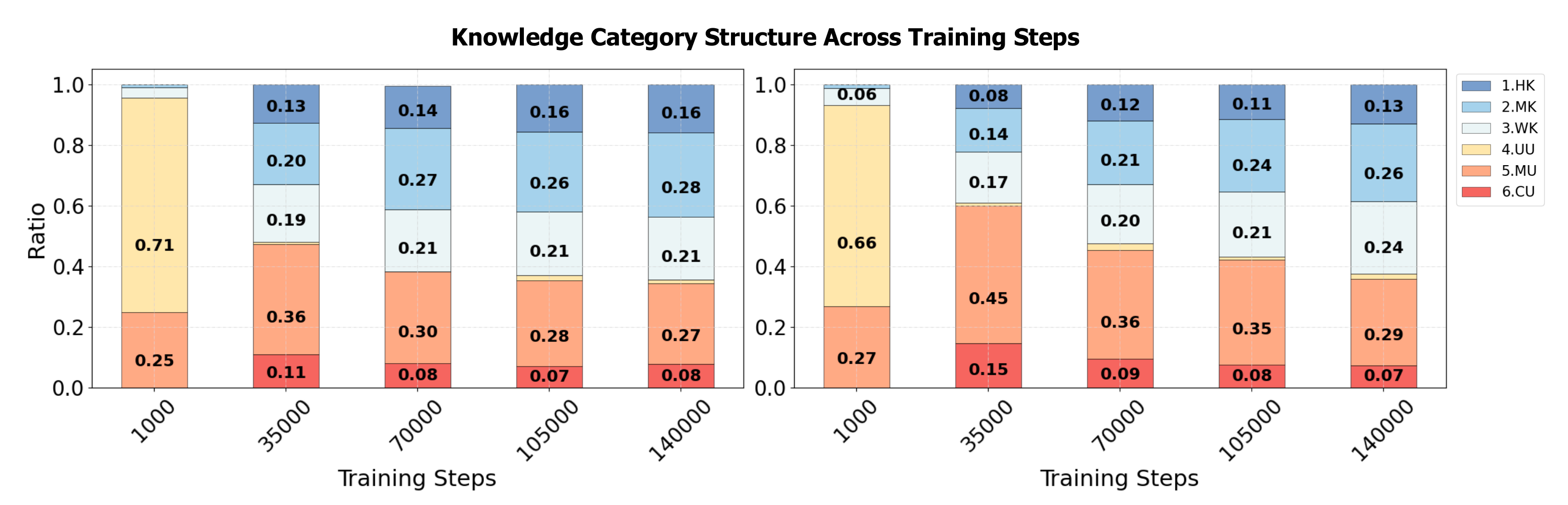}
   \caption{Changes in \textbf{external} knowledge categories across different training steps for the \textbf{Pythia-2.8b} (\textbf{Left} figure) and \textbf{Pythia-1.4b} (\textbf{Right} figure) model.}
    \label{fig:external_training_stesp_knowledge_structure}
\end{figure}

\begin{figure}[t]
    \centering
    
    \begin{subfigure}{0.493\linewidth}
        \centering
        \includegraphics[width=1.7in]{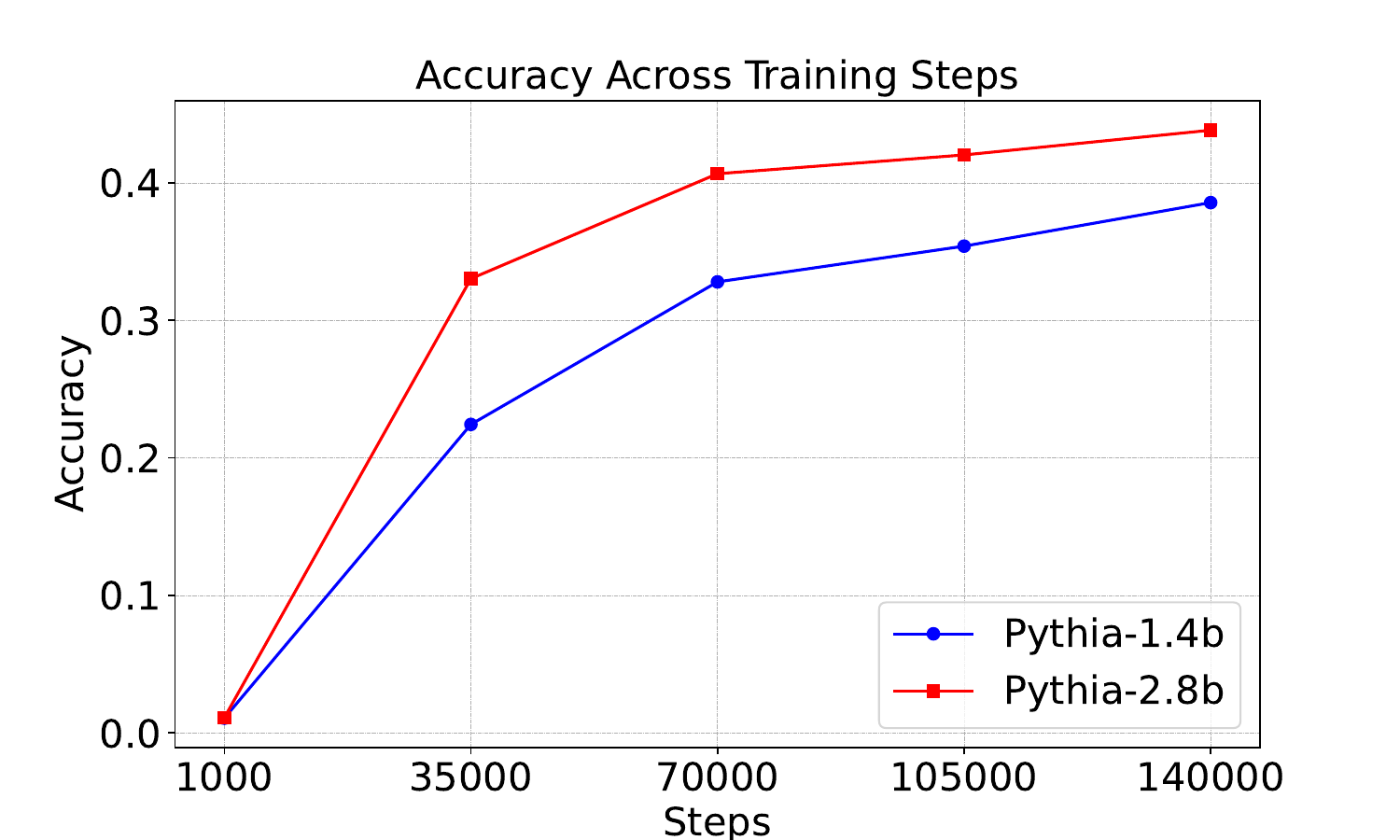}
        \caption{Accuracy}
    \end{subfigure}
    \begin{subfigure}{0.493\linewidth}
        \centering
        \includegraphics[width=1.7in]{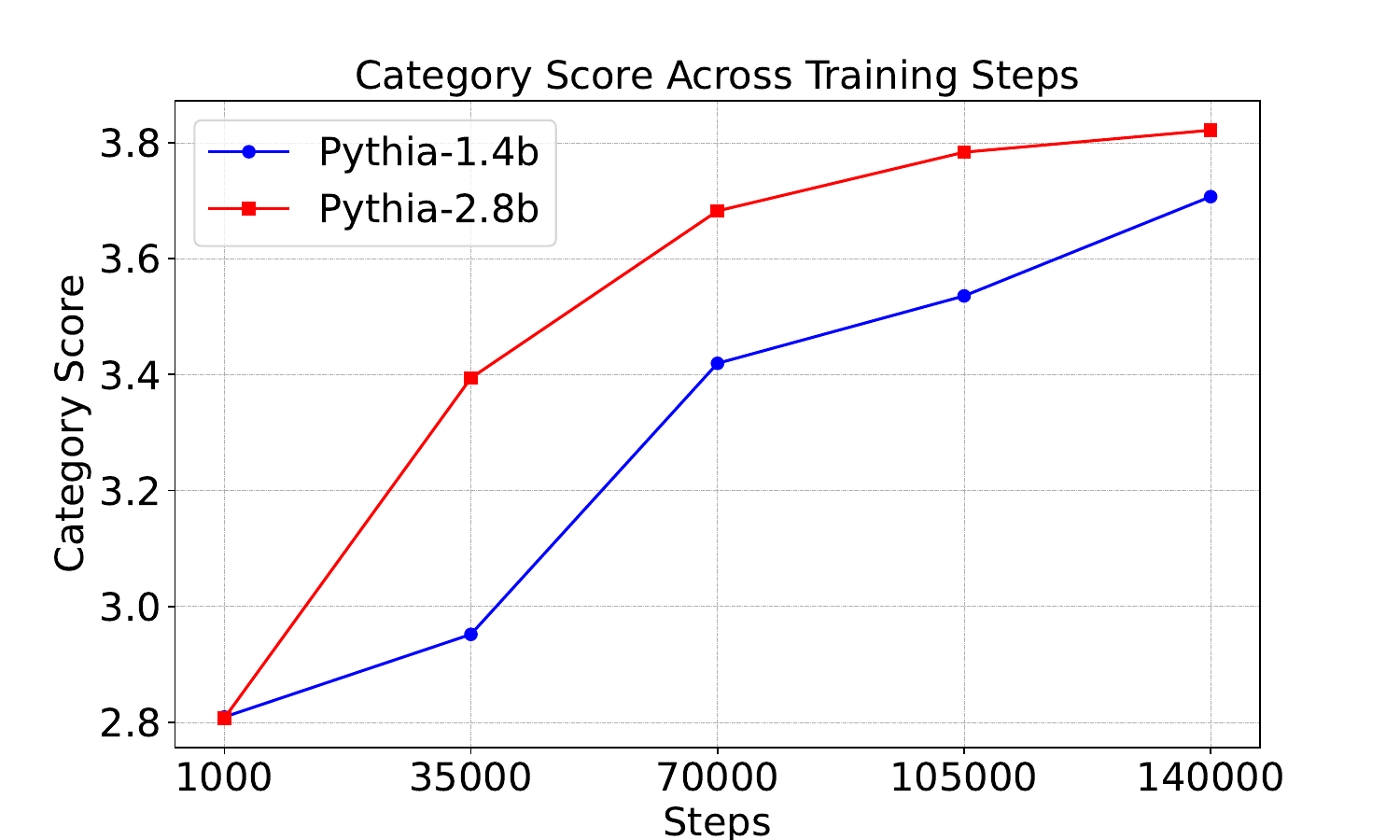}
        \caption{Category Score}
    \end{subfigure}
    \caption{Accuracy and Category Score for external knowledge during different training Steps.}
    \label{fig:external_accuracy_and_category_score}
\end{figure}

\begin{graybox}
    \textbf{Finding 8}: Models show more consistent learning patterns with provided context, though still develop conflicting knowledge representations.
\end{graybox} 
The model's changes, illustrated in Figure \ref{fig:external_training_stesp_knowledge_structure}, resemble those in knowledge category structure displayed in Figure \ref{fig:example_figure}. Models improve knowledge understanding, with knowledge category 5.MU initially expanded before contracting. However, even with contextual knowledge provided, the models still appear to encode conflicting knowledge, evidenced by an increase in the ratio of 6.CU. In contrast, for context understanding, accuracy is closely instructed with category scores as shown in Figure \ref{fig:external_accuracy_and_category_score}. Unlike internal knowledge, there is no pattern of category scores decreasing and then increasing for external knowledge.

\section{Conclusion}
\label{sec:conclusion}

We present K-(CSA)$^2$, a framework that evaluates LLMs' knowledge through correctness and confidence metrics. Through experiments with various models and techniques like CoT and RLHF, we uncover novel patterns in how knowledge is organized and evolves during training. This framework provides systematic metrics for assessing LLMs' knowledge structures, benefiting future models development.
\section*{Limitation}
\label{sec:limitation}

While our framework provides novel insights into LLM knowledge evaluation, several limitations should be noted. Our method relies heavily on sampling to determine confidence levels, where the number of samples and temperature settings could affect categorization results. The boundaries between categories, especially for "Maybe Known" and "Weakly Known", can sometimes be ambiguous. While our current definitions provide practical distinctions, more rigorous theoretical foundations for these boundaries could be developed. Additionally, evaluating models through multiple sampling runs is computationally intensive, which may limit the framework's applicability in resource-constrained scenarios. Our evaluation primarily focuses on factual knowledge in question-answering contexts and may need adaptation for other types of knowledge or tasks. For external knowledge evaluation, the framework doesn't fully capture how models integrate multiple pieces of context or handle conflicting information. Future work could address these limitations by exploring sampling robustness, strengthening theoretical foundations, and extending to more complex scenarios.

\bibliography{custom}

\clearpage

\appendix

\section{Appendix}
\label{sec:appendix}

\subsection{Related Work}
\label{sec:related_work}

%[Talking about knowledge-probing, emergency]
\textbf{Knowledge Representation in LLMs}: Studies have shown that LLMs can store extensive factual knowledge in their weights \cite{allen-zhu2024physics, sun-etal-2024-head}. However, these models face challenges in knowledge generalization and reasoning \cite{wu2024clashevalquantifyingtugofwarllms, zhang2024knowhaluhallucinationdetectionmultiform}. While traditional methods like knowledge probing \cite{kglens-towards-efficient} and question answering \cite{burns2024discoveringlatentknowledgelanguage} measure knowledge retention, they rarely examine the internal structure of model knowledge. Our framework extends beyond correctness assessment to analyze knowledge confidence and distribution across categories.

%[self-consistency, Context-aware Decoding]
% \noindent \textbf{LLM Inference Sampling:} Sampling-based techniques, such as random sampling during response generation, have been widely used to assess the flexibility and depth of knowledge encoded in LLMs. Previous works, such as \cite{wang2023selfconsistency, nguyen2024consistentpredictionlikelycorrect, manakul2023selfcheckgpt}, have explored introducing variability in LLM responses. \cite{shi-etal-2024-trusting} propose using contrastive decoding, where LLMs are prompted to answer questions both with and without contextual knowledge, aiming to reduce hallucinations. These methods have been instrumental in generating diverse responses for tasks like creative writing and open-ended question answering. Nevertheless, few approaches \cite{Gekhman2024DoesFL} systematically evaluate the implications of this variability for knowledge comprehension. In contrast, our framework employs sampling to examine the confidence of model outputs over repeated prompts, categorizing knowledge based on both accuracy and response diversity.

\noindent \textbf{Uncertainty in LLMs}: Research on LLM uncertainty \cite{kalai2024calibratedlanguagemodelshallucinate, yin-etal-2023-large} has revealed that models can generate overconfident incorrect answers \cite{azaria-mitchell-2023-internal, li2024thinktwicetrustingselfdetection}. Studies have explored various uncertainty quantification methods, including semantic uncertainty \cite{kuhn2023semantic} and estimation algorithms \cite{huang2023lookleapexploratorystudy}. Our framework uniquely combines correctness and confidence metrics by categorizing knowledge based on response consistency, providing deeper insights into model comprehension.

\subsection{Knowledge Category Detailed Description}

As shown in Table \ref{tab:knowledge_category_description}, the framework defines six distinct categories of knowledge comprehension based on model response patterns across greedy decoding and sampling approaches. Highly Known (1.HK) represents the strongest form of knowledge comprehension, where responses are both consistent and correct across all decoding methods. Maybe Known (2.MK) indicates partial knowledge with correct greedy decoding but inconsistent sampling results, while Weakly Known (3.WK) shows minimal knowledge through occasional correct sampling despite incorrect greedy decoding. The framework also identifies three categories of unknown knowledge: Unconfident Unknown (4.UU) where wrong answers vary with each sampling, May Confident Unknown (5.MU) with some repeated incorrect answers, and Confident Unknown (6.CU) representing consistent incorrect responses indicating strong misconceptions.

\begin{table}[h!]
    \centering
    \begin{tabular}{l|cccc}
        \hline
        \multirow{2}{*}{\textbf{Models}} & \multicolumn{4}{c}{\textbf{Analysis}}\\ \cline{2-5} 
        &KS&CoT&IT&TS \\ \hline
        GPT-4o mini &\checkmark &\checkmark & & \\
        \small\cite{openai2024gpt4ocard} &&&& \\ \hline
        GPT-3.5 &\checkmark & & & \\
        &&&&\\ \hline
        Qwen2-0.5b &\checkmark &\checkmark &\checkmark &\checkmark \\
        \small\cite{yang2024qwen2technicalreport}&&&& \\ \hline
        gemma-2-2b &\checkmark &\checkmark &\checkmark &\checkmark \\
        \small\cite{gemmateam2024gemma2improvingopen}&&&& \\ \hline
        Llama-2-7b &\checkmark &\checkmark &\checkmark &\checkmark \\
        \small\cite{touvron2023llama2openfoundation}&&&& \\ \hline
        Llama-3-8b &\checkmark & &\checkmark & \\
        \small\cite{grattafiori2024llama3herdmodels}&&&& \\ \hline
        Mistral-7b &\checkmark & &\checkmark & \\
        \small\cite{jiang2023mistral7b}&&&& \\ \hline
        Pythia-2.8b & & & &\checkmark \\
        \small\cite{biderman2023pythia}&&&& \\ \hline
        Pythia-1.4b & & & &\checkmark \\
        \small\cite{biderman2023pythia}&&&& \\ \hline
    \end{tabular}
    \caption{Collections of the models we used in our experiments and what sections they have been used to evaluate. \textbf{KS}: Knowledge Category Structure, \textbf{CoT}: Chain of Thoughts, \textbf{TS}: Training Steps, \textbf{IT}: Instructed model.}
    \label{table:model_collection}
\end{table}

\begin{table*}[h!]
    \centering
    \renewcommand{\arraystretch}{1.5} % Increase row height
    \scalebox{0.8}{\begin{tabular}{|c|l|c|l|}
        \hline
        \textbf{Category} & \textbf{Full name} & \textbf{Color} & \textbf{Description} \\ \hline
        1.HK & \textbf{Highly Known} & \colorbox[rgb]{0.294, 0.455, 0.698}{\strut\hspace{2em}} & Responses are consistent and correct \\ \hline
        2.MK & \textbf{Maybe Known} & \colorbox[rgb]{0.549, 0.745, 0.878}{\strut\hspace{2em}} & Response is correct in greedy decoding, but wrong in sampling. \\ \hline
        3.WK & \textbf{Weakly Known} & \colorbox[rgb]{0.902, 0.945, 0.953}{\strut\hspace{2em}} & Greedy decoding response is wrong, but there are few correct sampling answers. \\ \hline
        4.UU & \textbf{Unconfident Unknown} & \colorbox[rgb]{1.0, 0.875, 0.573}{\strut\hspace{2em}} & The model gives wrong answers, but answers are different at each sampling. \\ \hline
        5.MU & \textbf{May Confident Unknown} & \colorbox[rgb]{0.988, 0.549, 0.353}{\strut\hspace{2em}} & There are few repeated answers in sampled response, but not all. \\ \hline
        6.CU & \textbf{Confident Unknown} & \colorbox[rgb]{0.859, 0.192, 0.141}{\strut\hspace{2em}} & The model is fully confident but gives an incorrect answer consistently. \\ \hline
    \end{tabular}}
    \caption{Knowledge Category Table. The definitions of \textbf{Highly Known}, \textbf{Maybe Known}, \textbf{Weakly Known} are following \cite{Gekhman2024DoesFL}.}
    \label{tab:knowledge_category_description}
\end{table*}

\subsection{Confidence Calculation Illustration}
\label{subsec:confidence}

We calculate confidence by measuring the consistency of responses across multiple sampling iterations. Given a question $q$, context $a$, model $M$, and temperature $T$, the confidence score $P_{\text{Confidence}}(q, a; M, T)$ is computed as:
\begin{equation}
P_{\text{Confidence}}(q, a; M, T) = \max_{i} (\frac{f_i}{n})
\end{equation}
where $f_i$ represents the frequency of response $i$ and $n$ is the total number of samples. This metric is only computed for cases where $P_{\text{Correctness}}(q, a; M, T) = 0$, indicating that none of the sampled responses match the ground truth.
For instance, given the question "What is the largest planet in our solar system?" with ground truth "Jupiter", consider three scenarios with $n=5$ samples:
\begin{enumerate}
\item Complete uncertainty: responses ["Mars", "Venus", "Jupiter", "Saturn", "Neptune"] yield $P_{\text{Confidence}} = \frac{1}{5} = 0.2$, categorizing it as Unconfident Unknown (UU).
Copy\item Partial consistency: responses ["Saturn", "Mars", "Saturn", "Jupiter", "Mars"] result in $P_{\text{Confidence}} = \frac{2}{5} = 0.4$, placing it in the May confident Unknown (MU) category.

\item Complete consistency: responses ["Mars", "Mars", "Mars", "Mars", "Mars"] give $P_{\text{Confidence}} = \frac{5}{5} = 1.0$, indicating a Confident Unknown (CU) case.
\end{enumerate}
This confidence metric effectively distinguishes between three levels of uncertainty in incorrect responses: complete uncertainty (UU) where all responses differ, partial consistency (MU) where some responses match but without complete agreement, and complete consistency (CU) where all responses are identical but incorrect. For known categories (HK, MK, WK), where $P_{\text{Correctness}}(q, a; M, T) > 0$, confidence calculation is not applicable as the classification is determined by response accuracy rather than consistency.

\subsection{Example Of Using Category Score}
\label{sec:example_category_score}
For example, if a model produces responses for a set of questions that are 60\% 1.HK $(r_1 = 0.6)$ and 40\% 2.MK $(r_2 = 0.4)$, its Category Score would be $(6 \times 0.6) + (5 \times 0.4) = 5.6$, indicating strong knowledge comprehension. Conversely, model's responses concentrated in unkonwn categories would receive a much lower score, reflecting worse knowledge understanding.

\subsection{Further Explanations of Knowledge Transition Ratio}
\label{sec:explanaiton_ratios}
Importantly, these three ratios are mutually exclusive and satisfying that the sum of three metrics equals to $1$. Each metric captures a distinct aspect of knowledge evolution, allowing us to decompose changes in model behavior into independent components. For instance, a model might show high stability while having low upgrade and downgrade ratios, or exhibit significant changes with low stability but balanced upgrade and downgrade ratios. This decomposition provides a comprehensive view of how knowledge representation evolves during model development or across different variants.

For example, consider tracking these ratios across different scenarios:

\noindent\textbf{During fine-tuning on domain-specific data}: A model might show high upgrade ratios (0.5) for domain knowledge category, moderate stability (0.4) for general knowledge category, and low downgrades (0.1), indicating successful specialization without catastrophic forgetting. Alternatively, we might observe high downgrades in general knowledge category with high upgrades in domain knowledge category, suggesting a trade-off between specialization and general capabilities.

\noindent\textbf{Comparing model scale}: When moving from a smaller to larger model, we might see high upgrade ratios (0.6), moderate stability (0.3), and minimal downgrades (0.1), suggesting that scaling primarily improves capabilities. In contrast, moving to an extremely large model might show high stability (0.7) with balanced upgrades (0.15) and downgrades (0.15), indicating diminishing returns from scale.

\noindent\textbf{Analyzing iterative model improvements}: A version upgrade might display high stability (0.8) with small but positive upgrades (0.15) and minimal downgrades (0.05), suggesting controlled, conservative improvements. Another upgrade could show lower stability (0.4) but higher upgrades (0.45) and some downgrades (0.15), indicating more aggressive changes with both benefits and risks.
\begin{figure*}[h]
    \centering
    \begin{subfigure}{0.49\linewidth}
        \centering
        \includegraphics[width=\linewidth]{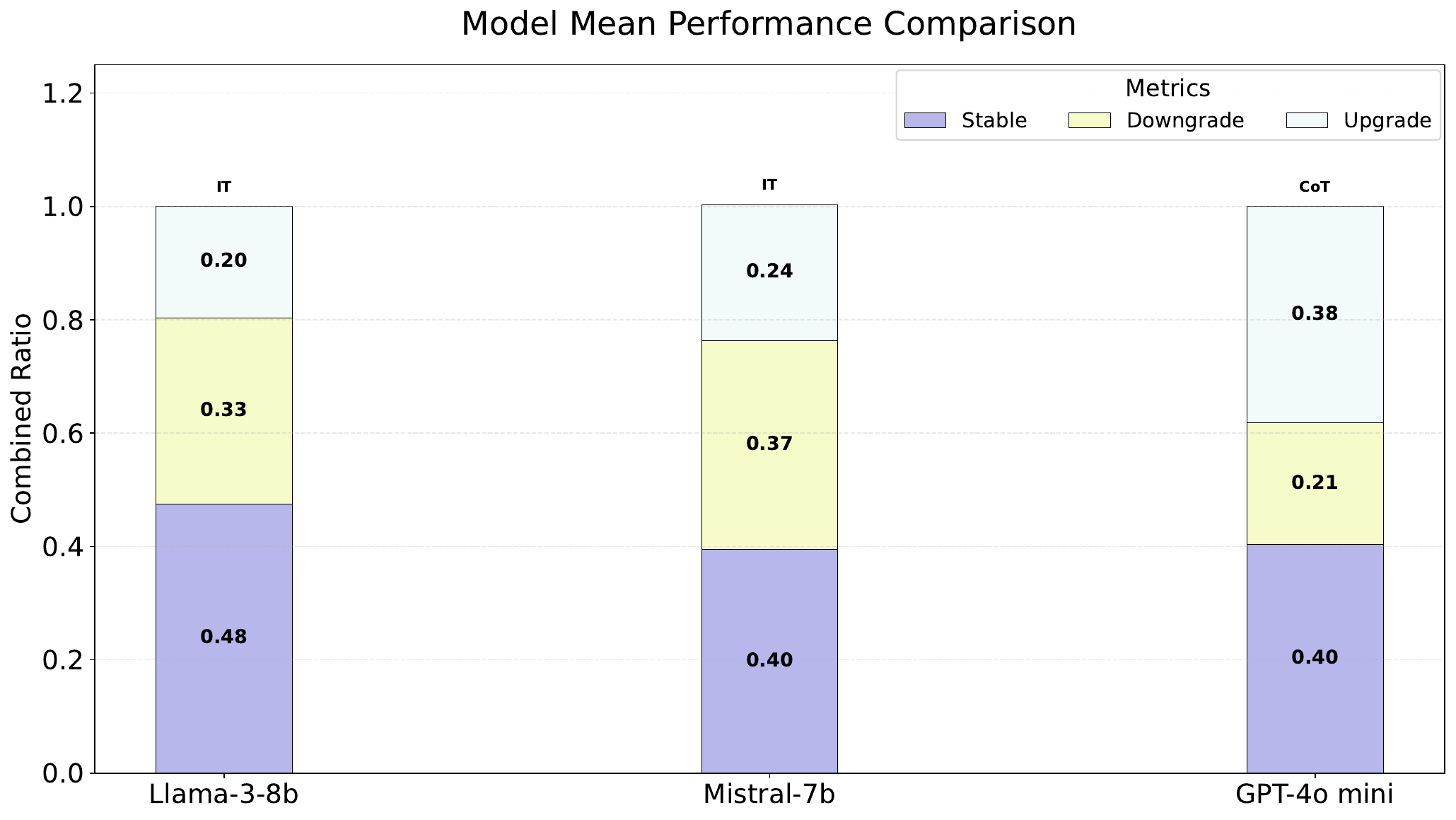}
        \caption{Internal Knowledge}
        \label{fig:internal_category_transition_it_model_mean}
    \end{subfigure}
    \begin{subfigure}{0.49\linewidth}
        \centering
        \includegraphics[width=\linewidth]{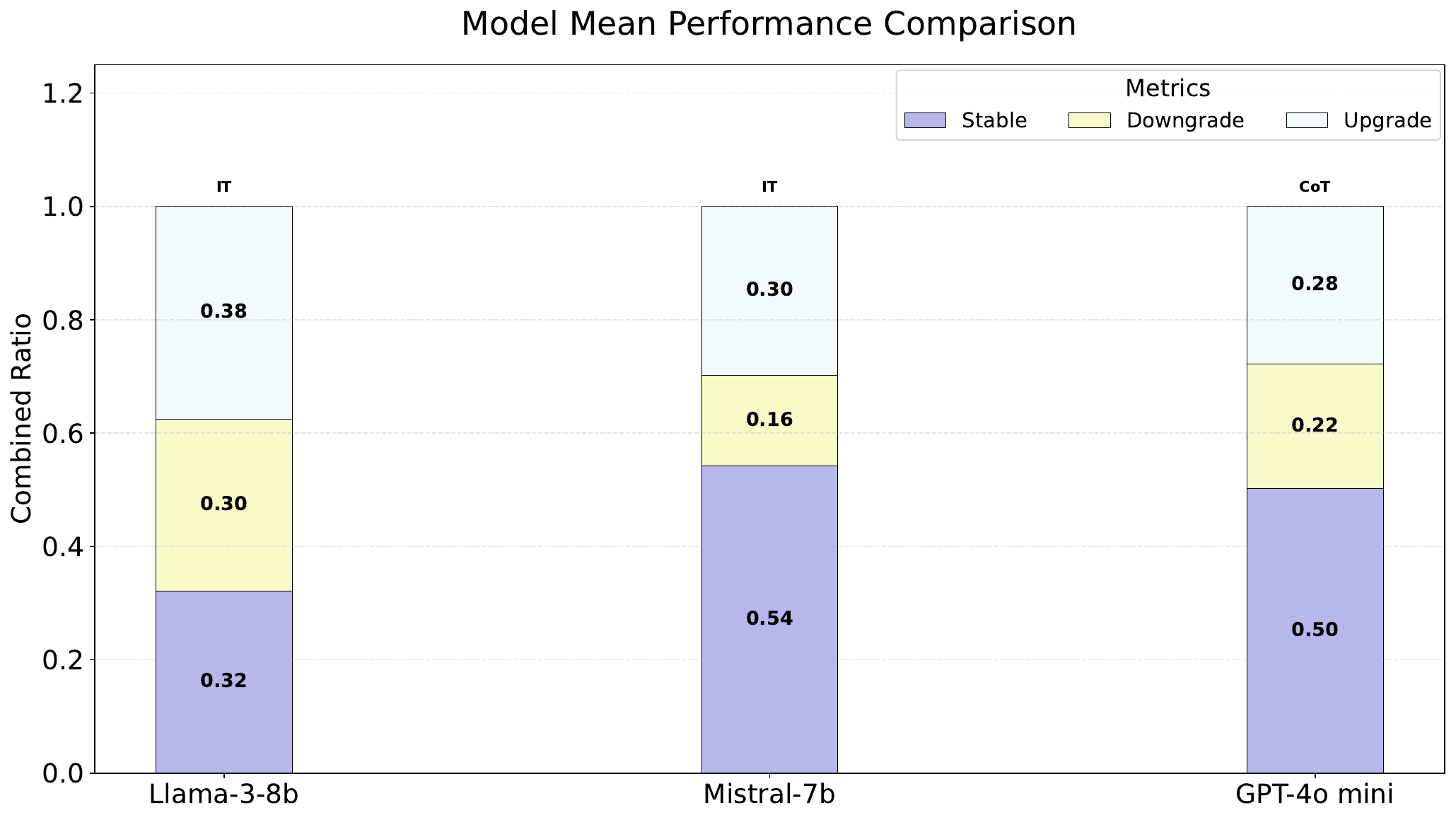}
        \caption{External Knowledge}
        \label{fig:others_model_performance}
    \end{subfigure}

    \caption{Additional results for knowledge category structure transition}
    \label{fig:category_transition_others}
\end{figure*}
\subsection{Models}
\label{subsec:model}

Table \ref{table:model_collection} is a collection of models and analysis involved. The experimental framework employs a diverse set of models analyzed across different dimensions. The core analysis of Knowledge Category Structure (KS) encompasses GPT-4o mini, GPT-3.5, Qwen2-0.5b, gemma-2-2b, Llama-2-7b, Llama-3-8b, and Mistral-7b. A subset of these models (GPT-4o mini, Qwen2-0.5b, gemma-2-2b, and Llama-2-7b) underwent Chain of Thought (CoT) analysis, while Instruction Tuning (IT) analysis was performed on Qwen2-0.5b, gemma-2-2b, Llama-2-7b, Llama-3-8b, and Mistral-7b. The Training Steps (TS) analysis specifically focused on Qwen2-0.5b, gemma-2-2b, Llama-2-7b, and included additional models Pythia-2.8b and Pythia-1.4b for comprehensive evaluation of training progression.

% \subsection{Sankey Diagram To Show Transition During Training}
% \label{subsec:sankey_training_steps}

\subsection{Additional Models' Knowledge Structure Transition Results}
\label{other_results}

Figure \ref{fig:category_transition_others} aims to display more results in additional to results presented in Figures 4 and 5.

\subsection{Detailed Transition Results Across Categories and Models}
\label{detailed_results}

\begin{figure*}[t]
    \centering
    \begin{subfigure}{0.9\linewidth}
        \centering
        \includegraphics[width=\linewidth]{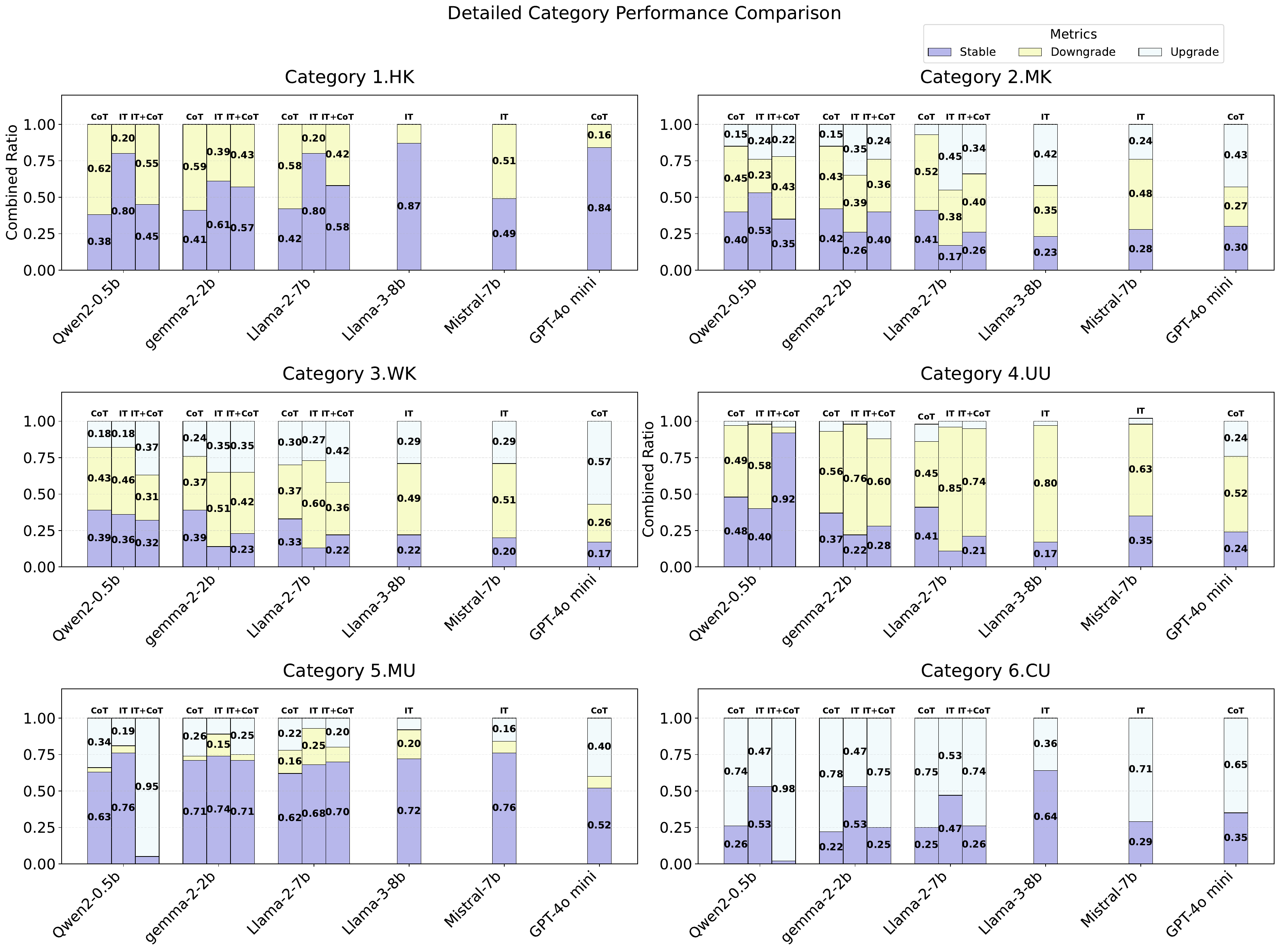}
        \caption{Internal Knowledge}
        \label{internal_detailed}
    \end{subfigure}\\
    \begin{subfigure}{0.9\linewidth}
        \centering
        \includegraphics[width=\linewidth]{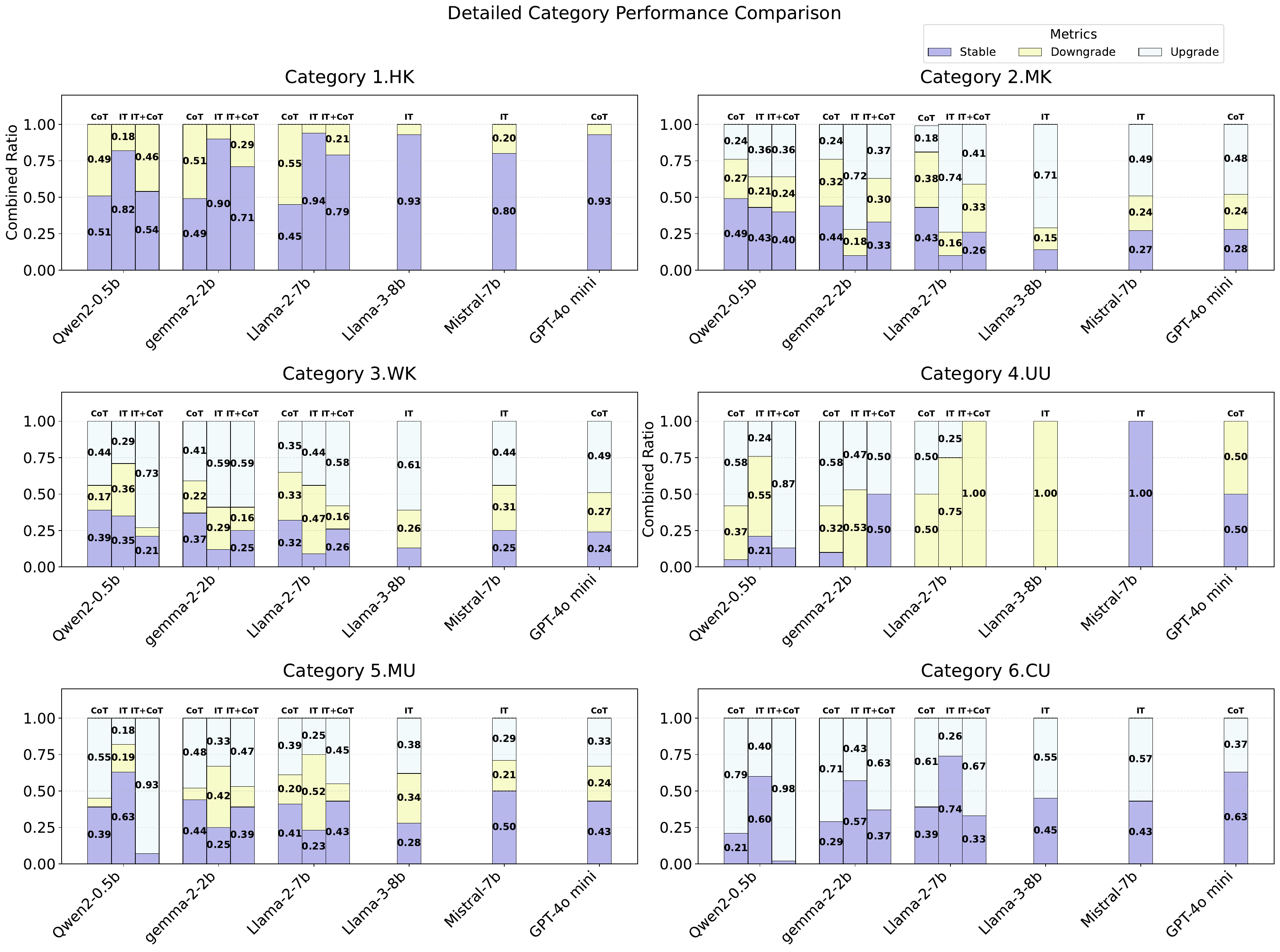}
        \caption{External Knowledge}
        \label{external_detailed}
    \end{subfigure}
    \caption{The detailed demonstrations of ratios for models across categories.}
    \label{fig:detail_ratio}
\end{figure*}

Figure \ref{fig:detail_ratio} is the comprehensive analysis of transition patterns reveals that instruction tuning and chain-of-thought prompting produce complementary effects across all knowledge categories. Models consistently show improved performance in higher-order categories (1.HK and 2.MK) when both techniques are applied, with particularly strong gains in the handling of external knowledge. The transition matrices demonstrate that knowledge tends to move upward through the category hierarchy during training, with the most significant improvements observed in the reduction of confident misconceptions (6.CU) and the strengthening of highly known information (1.HK).

% \begin{figure*}[t]
%     \centering
%     \includegraphics[width=1.0\linewidth]{figure/Introduction/illustration_2.pdf}
%     % \includegraphics[width=1\linewidth]{figure/Introduction/illustration_2.pdf}
%     \caption{Sankey Diagram to show the detail of knowledge category structure changes over training for model \textbf{Pythia-2.8b}}
%     \label{fig:introduction}
% \end{figure*}

\end{document}